%% file: main.tex
\documentclass{article}

\usepackage[final]{corl_2020} 

\usepackage[dvipsnames]{xcolor}
\usepackage{graphics}
\usepackage{graphicx}
\usepackage{amsthm}
\usepackage{subfigure}
\usepackage{amsmath,amssymb,mathtools}
\usepackage{amsfonts}  
\usepackage[ruled]{algorithm2e}
\usepackage[font={footnotesize}]{caption}
\usepackage{array}
\usepackage[utf8]{inputenc}
\usepackage{amsmath, amsthm, amssymb}
\usepackage{booktabs}
\usepackage{dsfont}
\usepackage{xcolor}
\usepackage{url}
\usepackage{enumitem}

\newcommand{\etal}{\textit{et al}.}

\usepackage{color, soul} 

\title{ROLL: Visual Self-Supervised Reinforcement Learning with Object Reasoning}

\author{
  Yufei Wang\thanks{Equal contribution},  ~Gautham Narayan Narasimhan\footnotemark[1], ~Xingyu Lin, ~Brian Okorn, ~David Held \\
  Carnegie Mellon University \\
  United States\\
  \texttt{\{yufeiw2, gauthamn, xlin3, bokorn, dheld\}@andrew.cmu.edu} 
}

\begin{document}
\maketitle

\input{inputs/def}

\begin{abstract}
    Current image-based reinforcement learning (RL) algorithms typically operate on the whole image without performing object-level reasoning.  This leads to inefficient goal sampling and ineffective reward functions. In this paper, we improve upon previous visual self-supervised RL by incorporating object-level reasoning and occlusion reasoning. Specifically, we use unknown object segmentation to ignore distractors in the scene for better reward computation and goal generation; we further enable occlusion reasoning by employing a novel auxiliary loss and training scheme. We demonstrate that our proposed algorithm,   \algo (Reinforcement learning with Object Level Learning), learns dramatically faster and achieves better final performance compared with previous methods in several simulated visual control tasks. Project video and code
are available at \url{https://sites.google.com/andrew.cmu.edu/roll}.
\end{abstract}




\input{inputs/1_intro}
\input{inputs/2_related-work}
\input{inputs/3_background}

\input{inputs/4_method}
\input{inputs/5_experiments}
\input{inputs/6_conclusion}



\clearpage
\acknowledgments{This material is based upon work supported by the United States Air Force and DARPA under Contract No. FA8750-18-C-0092, the National Science Foundation under Grant No. IIS-1849154, and LG Electronics.}


{
\bibliography{reference}  
}

\input{inputs/appendix}

\input{inputs/real-robot-generalization}

\end{document}

%% file: inputs/def.tex
\newcommand{\algo}{ROLL~}

%% file: inputs/1_intro.tex
\section{Introduction}
Recently, reinforcement learning (``RL'') has achieved great success in many robotics control tasks, such as pushing~\cite{agrawal2016learning, deepvisualforsight}, grasping~\cite{pinto2016supersizing, asymmetricAC}, navigation~\cite{pathak2018zero, lange2012autonomous}, and more.
One challenge for reinforcement learning in the real world is obtaining a reward function.
In order to learn directly in the real world from high-dimensional images, recent methods proposed to use a
learned reward function~\cite{nair2018visual, skewfit} using distance in a learned latent space. These methods also sample goals from this learned latent space for self-training on a diverse set of goals.

However, when learning the latent representation, most of these previous image-based self-supervised RL algorithms operate on the whole image without performing object-level reasoning.  This leads to inefficient goal sampling and an ineffective reward function. For example, in previous methods, a robot arm can be a distractor for policy learning, because it is appears in the robot observations and is thereby encoded into the latent space; thus, the reward derived from that latent space will account for the position of the robot arm information. Ideally, for object manipulation tasks, the sampled goals  and reward  should relate to only the target objects, rather than the robot arm or static parts of the scene. 

In this paper, we improve upon previous image-based self-supervised RL by incorporating object-level reasoning. Specifically,
to help the robot reason more about the target objects, we propose to use unknown object segmentation to segment out the target object and remove all other distractors in the scene; we show that this can allow for better reward computation and goal sampling. 

However, with a naive segmentation, the latent representation might suffer from object occlusions; if the object is occluded, then the segmented image might be empty. To handle this case, we augment our method with occlusion reasoning by using a novel occlusion-aware loss function and training scheme. Our proposed algorithm can generally be used in any setting that has a static background, which is common for many robotics tasks. Furthermore, it works with any type of objects and does not require any prior information about the objects; thus our method can be applied for learning various object manipulation tasks.

To summarize, the main contributions of our work are:
\begin{itemize}
    \item We propose a new algorithm, \algo (Reinforcement learning with Object Level Learning), that incorporates object-level reasoning to improve reward computation and goal sampling for visual self-supervised RL.
    \item We make \algo robust to object occlusions by using a novel auxiliary loss and training scheme.
    \item We demonstrate that \algo learns dramatically faster and achieves better final performance compared with previous methods in several simulated visual control tasks.
\end{itemize}

%% file: inputs/2_related-work.tex
\section{Related Work}
\textbf{Image-based RL:}
Image-based deep reinforcement learning has been applied to robotics tasks to learn a
variety of behaviors including navigation~\cite{lange2012autonomous}, grasping~\cite{asymmetricAC, bodnar2019quantile}, pushing~\cite{agrawal2016learning}, and more. Goal-conditioned RL has the potential of learning
general-purpose policies for performing different tasks specified as different goals~\cite{schaul2015universal, sutton2011horde, HER}.
Recent work in goal-conditioned image-based RL uses goals defined as images~\cite{skewfit, nair2018visual, lin2019reinforcement, nair2020contextual}. These approaches learn a latent encoding of the image goals and then condition the policy on the low-dimensional encoded goal vector. However, such methods typically also encode information that are not part of the intended goal, such as the position of the robot arm. Our work also uses image goals, but we perform object-level reasoning, leading to more robust goal representations and faster learning.

\textbf{Object reasoning in reinforcement learning: }
Object-oriented RL and OO-MDP~\cite{diuk2008object} have been proposed as alternatives to classic MDPs to improve data efficiency and generalization by leveraging representations of objects and their relations. Most of the existing approaches
require hand-crafted object representations and their relations~\cite{cobo2013object, diuk2008object, garnelo2016towards}, with some recent work aiming to automatically discover these from data~\cite{watters2019cobra,keramati2018strategic,zambaldi2018relational,veerapaneni2020entity}. 
In a recent work~\cite{keramati2018strategic}, object-oriented RL is used to help exploration. Relational RL~\cite{zambaldi2018relational} uses an attention mechanism to extract the relations between objects in the scene to help policy learning. 
In our method, we use unknown object segmentation to extract the object representation directly from images, which does not need any prior knowledge of the object. Similar to  previous work, we also use object reasoning to improve the data efficiency for RL learning, but we do so to obtain a better reward function in the self-supervised RL setting. 

\textbf{Occlusion reasoning in robotics: } 
Object occlusions pose a huge challenge for robot learning: even simple distractors that occasionally occlude the object can cause state-of-the-art RL algorithms to fail~\cite{cheng2018reinforcement}. Theoretically, occlusions can be modeled using the framework of POMDP~\cite{kaelbling1998planning}, but usually such a formulation is intractable to solve, especially when the states are image observations. Other than POMDP, there has been lots of work aiming to solve the occlusion problem. Cheng, \etal~\cite{cheng2018reinforcement} use active vision which  learns a policy to  move the camera to avoid occlusions. To track the possibly occluded pixels, Ebert, \etal~\cite{ebert2017self} use a Conv-LSTM with temporal skip connections to copy pixels from prior images in the history. We also use an LSTM to handle occlusions, but we further use an occlusion-aware loss and training procedure to make it more robust.

%% file: inputs/3_background.tex
\section{Problem Formulation}
We consider the setting in which a robot needs to learn to achieve a parameterized set of  goals, e.g., pushing a puck to various target locations, with the goals specified using images. The robot only has RGB sensor inputs and has no access to the ground-truth states, e.g., puck positions. Formally, this can be formulated as image-based goal-conditioned RL, with a state space $S$ of images, a goal space $G$ which we assume to be the same as $S$, an action space $A$, a transition dynamics $T: S \times A \times S \rightarrow [0,1]$,  and a reward function $r: S \times A \times G \rightarrow  \mathcal{R}$. Given the current observation $I_t$ and the goal image $I_g$, the robot needs to learn a goal-conditioned policy $\pi(\cdot | I_t, I_g): S \times G \rightarrow [0,1]$ that maps the image $I_t \in S$ and goal $I_g \in G$ to a distribution over actions $a \in A$ to reach the goal. 

As it is hard to specify a reward function and learn a policy directly based on high-dimensional images~\cite{nair2018visual, lin2019reinforcement}, previous work~\cite{nair2018visual, skewfit} has employed a $\beta$-VAE~\cite{beta-vae}.  The $\beta$-VAE consists of an encoder $f_\phi(\cdot)$ to encode the image $I_t$ to a latent vector $z_t = f_\phi(I_t)$.  This previous work then uses $z_t$ as the input to the policy and for computing the reward. Specifically, a latent embedding is computed for both the observation image $z_t = f_\phi(I_t)$ and the goal image $z_g = f_\phi(I_g)$. The policy $\pi(\cdot | z_t, z_g)$ is conditioned on both of these latent embeddings.  
Further, because it is assumed that the robot does not have access to the ground-truth state of the environment, it must use its observation images to compute the reward.  In this past work, the reward is calculated as the negative L2 distance between the observation and goal latent embeddings, $r_t = -||z_t - z_g||_2^2$.

%% file: inputs/4_method.tex
\section{Method}
\begin{figure}
    \centering
    \includegraphics[width=0.76\textwidth]{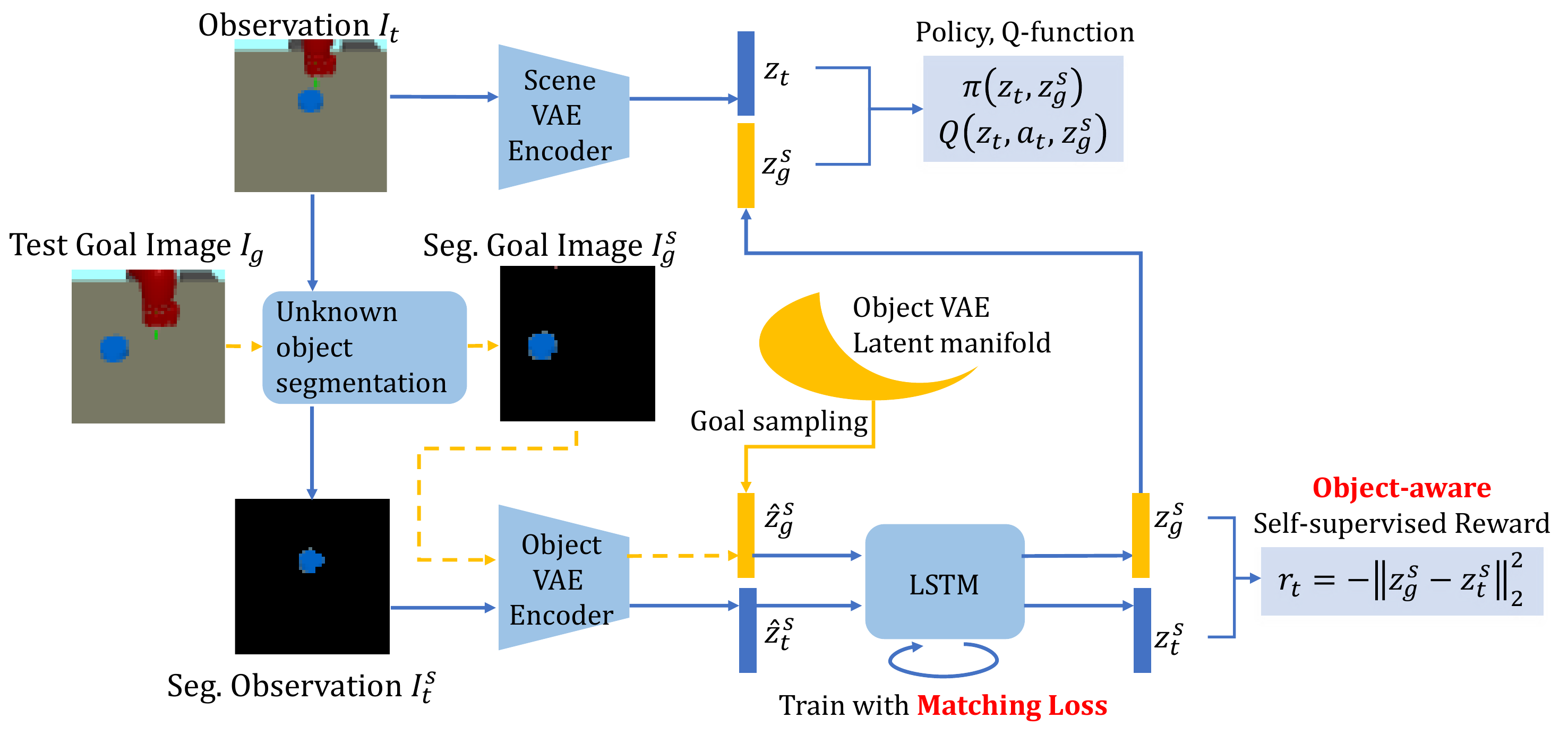}
    \caption{Overview of our proposed method. Top part: a scene-VAE encodes the whole image observation $I_t$ to the latent vector $z_t$, which is used as input to the policy and Q-function. Bottom part: an object-VAE encodes the segmented image $I_t^s$ to the latent vector $\hat{z}_t^s$, which is further encoded using an LSTM and used for object-aware reward computation and goal-conditioning. We train the LSTM with a novel matching loss and training scheme to make it robust to occlusions.
    At training time, the latent goal $z_g^s$ is obtained by sampling from the object-VAE prior and encoding by the LSTM. During test time, we segment the goal image and then encode it to the latent goal vector.}
    \label{fig:system}
\end{figure}

\subsection{Key Intuition and Overview}
The key intuition behind our method is that, for object manipulation tasks, goals are usually intended to refer to movable objects in the environment, rather than the robot arm or static parts of the scene. Our experiments show that, for previous methods, a robot arm can be a distractor for policy learning.  Specifically,  it distorts the reward function to account for the position of the robot arm information, whereas, in most cases, the intended reward function should ignore the robot arm and only consider the target object.  This error occurs because the VAE encodes the entire observation image into a latent vector, including the robot arm, and this VAE encoding is used for the reward computation.  

Therefore, to help the robot reason more about the target objects, we propose unknown object segmentation to segment out the target object and remove all other distractors in the scene for better reward computation and goal sampling. However, we found that, if we naively segment out the target object from the scene, then our method performs poorly when the object is occluded; hence, we use an LSTM and a novel occlusion-aware loss function and training procedure to train our method to be robust to occlusions. An overview of our method is shown in Figure \ref{fig:system}; in the following subsections, we detail how we perform unknown object segmentation and use segmented images for reward computation and how we use a novel auxiliary loss to make our method robust to  occlusions.

\subsection{Unknown Object Segmentation}
\label{sec:unkown_object_segmentation}
\begin{figure}
\centering
\begin{tabular}{ccccc}
     \includegraphics[width=0.1\textwidth]{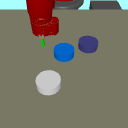}&  
     \includegraphics[width=0.1\textwidth]{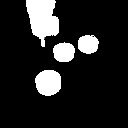}&
     \includegraphics[width=0.1\textwidth]{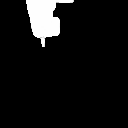}&
     \includegraphics[width=0.1\textwidth]{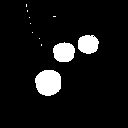}&
     \includegraphics[width=0.1\textwidth]{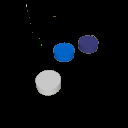} \\
     (a) original image  & (b) foreground mask  & (c) robot mask & (d) object mask & (e) segmented objects
\end{tabular}
\caption{ An illustration of the unknown object segmentation process. The method works with various number of objects with different properties and does not require prior information about them. We do not evaluate ROLL on this environment; it's just used to visualize the segmentation process. Here we visualize the: (a) original image; (b) foreground mask obtained by background subtraction; (c) robot mask predicted by robot segmentation; (d) object mask obtained by subtracting the robot mask from the foreground mask (thresholded at 0 to remove negative values); (e) final segmented objects. Note that the unknown object segmentation process does not explicitly reason about independent object instances or object relations.}
\label{fig:segmentation_process}
\end{figure}

Our method for unknown object segmentation  requires no prior information of the target object beforehand. Instead, it works by removing the pixels of everything else but the object from the scene. We note that other methods for unknown object segmentation could be used here~\cite{xie2020best,xie2020unseen,xiang2020learning}; the method we use was chosen for its robustness and simplicity.

We assume that the background is static, which is often the case for robot manipulation tasks. At training time, our first goal is to learn a model to segment out the robot from the scene.  To do so, we make the robot move with random actions, and we record images during this movement.  We then train a background subtraction module on this scene~\cite{GMM_bg_subtraction1, GMM_bg_subtraction2,opencv_library} (see Appendix for details). Because the robot is the only moving object in the scene during this data collection, it will be the only object not included as part of the background.  We use this  background subtraction module to generate ``ground-truth" segmentation labels
to segment the robot from the background; we then use these labels to train a robot segmentation network (see Appendix for details).

At test time, we add an unknown target object into the scene for robot manipulation.  The background subtraction module removes the static background, leaving the pixels for the robot and the newly placed objects.  The robot segmentation network is used to remove the robot pixels, leaving only the object pixels.
The full test-time segmentation process is shown in Figure~\ref{fig:segmentation_process}. As noted above, other methods could also be used for unknown object segmentation~\cite{xie2020best,xie2020unseen,xiang2020learning}.

\subsection{Using Segmented Images for Reward Computation}

We next describe how we can use these segmented images for a more robust reward computation.  See Figure~\ref{fig:system} for an overview of our system.  Our method employs two encoders: given the current observation $I_t$, a ``scene-VAE" 
encoder $f_\phi$ operates on $I_t$ and encodes it to a latent vector $z_t = f_\phi(I_t)$ that is used as the policy / Q-function input  (top part of Figure~\ref{fig:system}).
The ``object-VAE" 
$f_{\phi^s}$ operates  on the segmented image $I_t^s$ (bottom part of Figure~\ref{fig:system}), obtained through the above method for unknown object segmentation which creates an image that contains only the objects.  The object-VAE encodes the segmented image into another latent vector $\hat{z}_t^s = f_{\phi^s}(I_t^s)$.  Occlusions can cause the object to not appear in the image, leading to an ``empty" segmented observation or goal image.  To overcome this, we augment our network with occlusion reasoning, using an LSTM as well as a novel occlusion-aware loss function (details in Section~\ref{sec:robust_to_occlusion}). The final object encoding $z_t^s$ is the output of the LSTM using the object-VAE latent $\hat{z}_t^s$ as its input. A similar procedure is applied to the goal image $I_g$ to obtain an encoding $z_g^s$ of the segmented goal image (see Figure~\ref{fig:system}).

The object-VAE is used by our method in four ways for reward computation and goal conditioning.  First, the reward is computed as the negative L2 distance between the object latent encoding of the segmented observation and goal images: $r_t = -||z_t^s - z_g^s||_2^2$.  Second, the policy and Q-function take the scene-VAE latent encoding $z_t$ as input, but they are both goal-conditioned on the object latent encoding $z_g^s$.  
Third, during training, the goal object latent encoding is sampled from the object-VAE latent manifold and encoded using the LSTM. Last, at test time, the test goal image is first segmented and then encoded using the object-VAE and the LSTM to obtain the goal object latent encoding.

By employing unknown object segmentation and using the segmented object latent encoding for reward computation and goal conditioning, we ensure that only the object information is used for guiding the learning of the policy, allowing the policy to ignore the the robot arm and the background for reward computation and goal conditioning. Our experimental results demonstrate that using the segmented object latent encoding dramatically speeds up learning in many environments.

\subsection{Robustness to Object Occlusions}
\label{sec:robust_to_occlusion}

As our reward computation and goal sampling depends solely on the segmented target object, occlusions of the objects become an issue of our method. When an object is largely or totally occluded, the segmented image contains only a few (or zero) pixels, making the corresponding latent encoding non-informative for reward computation. Furthermore, if the object-VAE is trained on many images with large occlusions, many of the goals sampled from its latent manifold will also contain large occlusions, making them improper goals for policy training. 
An example of the robot arm occluding the puck is shown in Figure \ref{fig:environment_imgs}(c). In this section, we describe how we use an LSTM with a novel ``matching loss" and occlusion-aware training procedure, which allows the robot to implicitly estimate the object's current position by encoding the history of the object's previous positions.

Our intuition for using an LSTM~\cite{lstm} to estimate the object's current position under occlusion is based on the following two observations.  First, objects usually move in a smooth motion.  Second, during object manipulation, an occlusion (especially by the robot arm) usually lasts for just a small number of frames. Therefore, it is feasible that the object's position in the occluded frames can be inferred from the previous positions of the object, which are encoded in the LSTM. Specifically, the object-VAE latent vector $\hat{z}_t^s$ is input to the LSTM to obtain the final object latent encoding as $z_t^s = g_\psi(\hat{z}_t^s, h_{t-1})$, where $h_{t-1}$ is the hidden state from the previous time step and $g_\psi$ is the LSTM encoding with parameters $\psi$.

We use self-supervised losses to reduce the number of samples needed to train the LSTM.  First, we train the LSTM with an auto-encoder loss to reconstruct its input: $L^{ae}(\psi) = || z_t^s - \hat{z}_t^s ||_2^2$. To further make the latent encoding $z_t^s$ robust to occlusions, we propose a novel \emph{matching loss}: given a trajectory of segmented images $I^s_1, ..., I^s_t, ..., I^s_T$, we first use the LSTM to encode them as $z^s_1, ..., z^s_t, ..., z_T^s$. Next, we randomly pick an image in the trajectory $I_t^s$ with $t > 1$, to which we add  synthetic occlusions to obtain $I^{s, occ}_t$.  Since $I_t^s$ is a segmented object image (see Figure~\ref{fig:segmentation_process}(e)), adding a synthetic occlusion is achieved by removing additional pixels from the segmented object.  Then, we create a new ``occluded" trajectory, in which we replace $I_t^s$ by the occluded image  $I^{s, occ}_t$; thus, the new trajectory is given by $I^s_1, ..., I^{s, occ}_t, ..., I^s_T$.  We use the LSTM to encode this new occluded trajectory to obtain the new encodings $z_1^s, ..., z^{s, *}_t, ..., z^{s, *}_T$. Note that the encodings after time step $t$ will change since the input at time $t$ has changed and since the LSTM also encodes the history of previous inputs.

The matching loss is designed to ensure that the encodings of the occluded images, and the images thereafter, are the same as the encodings of the unoccluded images.  This loss is computed as: 
\begin{equation}
L^{matching}(\psi) = \frac{1}{T-t+1} \sum_{i=t}^T ||z_i^s - z_i^{s, *} ||^2_2
\end{equation}
In other words, we force the LSTM to encode every image after the manual occlusion in the occlusion trajectory to be the same as its counterpart in the original trajectory.  Specifically, for timestep $t$ in which the input $I^s_t$ was replaced by an occluded input image $I^{s, occ}_t$, the encodings $z_t^s$ and $z_t^{s, *}$ are enforced to be the same.  Thus, the LSTM must use the history of the images before timestep $t$ to estimate the position of the object at timestep $t$ in which it is occluded.  

We also add a similar matching loss to the latent encoding of the object-VAE: given an image $I_t^s$, we add manual occlusions to it to obtain $I_t^{s, occ}$, and we require the object-VAE to encode these images as closely as possible:
\begin{equation}
L^{matching}(\phi^s) = ||f_{\phi^s}(I_t^s) - f_{\phi^s}(I_t^{s, occ}) ||^2_2
\end{equation}
Though the object-VAE is before the LSTM and has no history information, we still find that this loss leads to more stable training when combined with the matching loss on the LSTM output.  In our experiments, we show that this novel matching loss allows our method to be robust to occlusions.

\subsection{Algorithm Summary}
We now explain how we train each part of our method. For the scene-VAE, we first pretrain it on a dataset of images collected using a random policy. We also continue training it using images stored in the replay buffer during RL policy learning. The scene-VAE is trained using the standard $\beta$-VAE losses, i.e., image reconstruction loss and the KL regularization loss in the latent space. For the object-VAE, we also pretrain it on a dataset of segmented images, where we randomly move the object in the scene. The object-VAE is trained using the regular $\beta$-VAE loss as well as the matching loss described in the previous section. After we pretrain the object-VAE, it is fixed during the RL training process. 
We pretrain the LSTM using the auto-encoder loss using the dataset of segmented images mentioned above for training the object-VAE; we also train it online using trajectories generated by the learning RL policy. This removes the burden to pre-collect a set of trajectories for training the LSTM. 
We train the LSTM online using the auto-encoder loss and matching loss as described in the previous section. For RL policy training, we use SAC~\cite{sac} with goal re-labeling as in Hindsight Experience Replay~\cite{HER}. Algorithm~\ref{algo:full} summarizes our method. We use blue text to mark the novel steps of our method compared to RIG~\cite{nair2018visual}. 

\begin{algorithm}
\SetAlgoLined
 \SetKwInOut{Input}{Input}
 \Input{scene-VAE $f_\phi$, object-VAE $f_{\phi^s}$, LSTM $g_\psi$, policy $\pi_\theta$, Q function $Q_w$. }
 
        \footnotesize
       
 Collect initial scene data $\mathcal D = \{I_{(i)}\}$ using random initial policy;
train scene-VAE $f_\phi$ on $\mathcal D$ \\
\textcolor{blue}{
Collect initial object data $\hat{\mathcal D} = \{\hat{I}_{(i)}\}$ by randomly placing the object in the scene \\
Use unknown object segmentation on $\hat{\mathcal D}$ to obtain segmented images $\mathcal D^s = \{I^s_{(i)}\}$ ; Train object-VAE $f_{\phi^s}$ on  $\mathcal D^s$ with image reconstruction, KL regularization, and matching loss; Train LSTM $g_\psi$ on $\mathcal D^s$ with auto-encoder loss} \\

    \For{$n=0,...,N-1$ epochs}{
         \textcolor{blue}{Sample goal embedding from object-VAE prior $\hat{z}^s_g \sim p(\hat{z}^s)$ and encode using LSTM: $z^s_g = g_\psi(\hat{z}^s_g)$
         }
         \\

        \For{$t=0,...,T$ steps}{
             // Collect data \\
             Embed the observation $I_t$ with the scene-VAE  $f_\phi(I_t)$ \\
             \textcolor{blue}{Get action $a_t \sim \pi_\theta(f_\phi(I_t), z^s_g)$ (conditioned on object latent goal embedding)} \\
             Execute action $a_t$ to get next observation $I_{t+1}$;  
             store $(I_t, a_t, I_{t+1}, z^s_g)$ into replay buffer $\mathcal R$ \\
             // Train the policy \\
            Sample transition $(I, a, I', z^s_g) \sim \mathcal R$ \\
             Encode images with scene-VAE $z = f_{\phi}(I), z' = f_{\phi}(I')$ \\
             \textcolor{blue}{Perform unknown object segmentation on $I$ and $I'$ to obtain $I^s$, $I^s{}'$} \\
             \textcolor{blue}{Encode segmented images with object-VAE and LSTM $z^s = g_\psi(f_{\phi^s}(I^s)), z^s{}' = g_\psi( f_{\phi^s}(I^s{}'))$} \\
             Hindsight Experience Replay (RIG version): With probability 0.5, replace $z^s_g$ with $z^s_g{}'$, \textcolor{blue}{which is sampled from object-VAE prior $\hat{z}^s_g{}' \sim p(\hat{z}^s)$ and embedded with LSTM: $z_g^s{}' = g_\psi(\hat{z}^s_g{}')$} \\
             \textcolor{blue}{Compute reward as $r = -||z^s{}' - z^s_g||_2^2$ (distance between object latent embeddings for observation and goal)}\\
             Train $\pi_\theta$, $Q_w$ using scene-VAE embeddings $z, z'$ \textcolor{blue}{and object latent goal embedding $z_g^s$}: $(z, a, z', \textcolor{blue}{z^s_g}, r)$.
             
        }
        Perform Hindsight Experience Replay with future observations \\
     
         Train scene-VAE $f_\phi$ with  image reconstruction loss and KL regularization loss \\ 
         \textcolor{blue}{Train LSTM $g_\psi$ with auto-encoder loss and matching loss  \\
         }
    }
    
 \caption{ROLL: Visual Self-supervised RL with Object Reasoning}
 \label{algo:full}
\end{algorithm}

%% file: inputs/5_experiments.tex
\section{Experiments}
In our experiments, we seek answers to the following questions: (1) Does using unknown object segmentation provides a better reward function and improve upon the baseline? (2) Does using an LSTM trained with the matching loss make \algo robust to occlusions?

\subsection{Setups}
\begin{figure}
    \centering
    \begin{tabular}{ccccc}
        \includegraphics[width=0.11\textwidth]{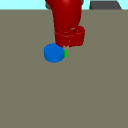} &
        \includegraphics[width=0.11\textwidth]{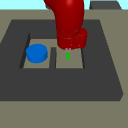}& 
        \includegraphics[width=0.11\textwidth]{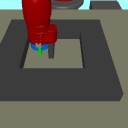} &
        \includegraphics[width=0.11\textwidth]{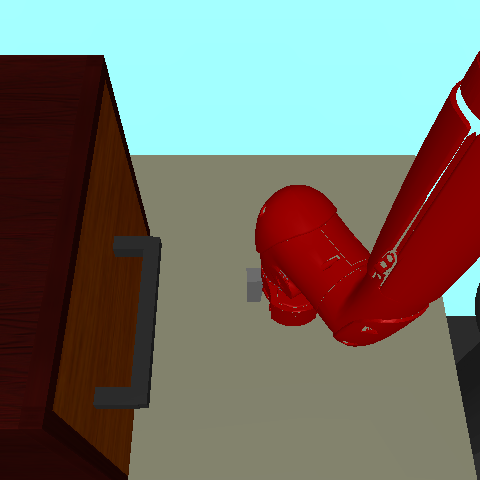}&
        \includegraphics[width=0.11\textwidth]{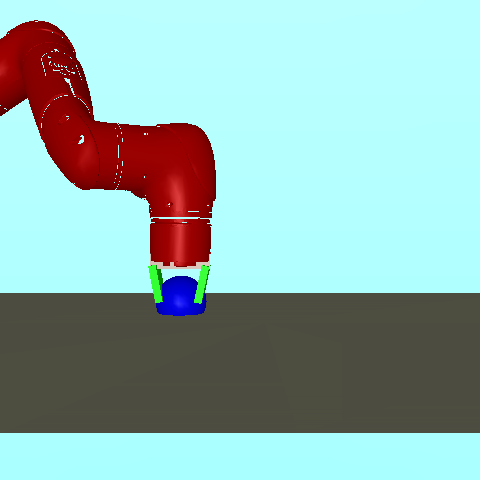} \\

        (a)  & (b) & (c)  &(d) & (e) 
    \end{tabular}
    \caption{ The robot view of different tasks: (a) Puck Pushing (b) Hurdle-Bottom Puck Pushing (c) Hurdle-Top Puck Pushing (d) Door Opening (e) Object Pickup. 
    }
    \label{fig:environment_imgs}
\end{figure}

We evaluate \algo on five image-based continuous control tasks simulated using the MuJoCo~\cite{todorov2012mujoco} physics simulator, where the policy must learn to manipulate objects to achieve various goals using only images as policy inputs, without any state-based or task-specific
reward. The goals are specified using images. The robot views of different tasks are illustrated in Figure~\ref{fig:environment_imgs}. 
\textbf{Puck Pushing:} a Sawyer robot arm must learn to push a puck to various goal locations. 
\textbf{Hurdle-Bottom Puck Pushing:} this task is similar to puck pushing, where we add hurdles. The robot arm and puck are initialized to the top right corner and the arm needs to push the puck to various locations in the left column.
\textbf{Hurdle-Top Puck Pushing:} this task is the same as Hurdle-Bottom Puck Pushing except that the position of the hurdle is flipped. This poses the challenge of pushing the puck under occlusions, as the arm will largely occlude the puck when it pushes it from bottom to top (see Figure~\ref{fig:environment_imgs}(c)). 
\textbf{Door Opening:} a Sawyer arm must learn to open the door to different goal angles.
\textbf{Object Pickup:} a Sawyer arm must learn to pick up a ball object to various goal locations. More details on the environments are provided in the appendix. 

We compare our method to Skew-Fit~\cite{skewfit}, a state-of-art visual self-supervised goal-conditioned learning algorithm which uses a single scene-VAE for both policy input and reward computation. Our method extends Skew-Fit by using an object-VAE to ignore distractors and by adding a matching loss for occlusion reasoning. 
In Skew-Fit, they do not pretrain the $\beta$-VAE on a set of images. As \algo requires pre-training, for a fair comparison, we also pretrain the $\beta$-VAE in Skew-Fit using the same dataset we collect for pre-training the scene-VAE in ROLL. Implementation details and the full list of hyper-parameters and network architectures can be found in the appendix.

\subsection{Results}

\begin{figure}
    \centering
    \includegraphics[width=0.315\textwidth]{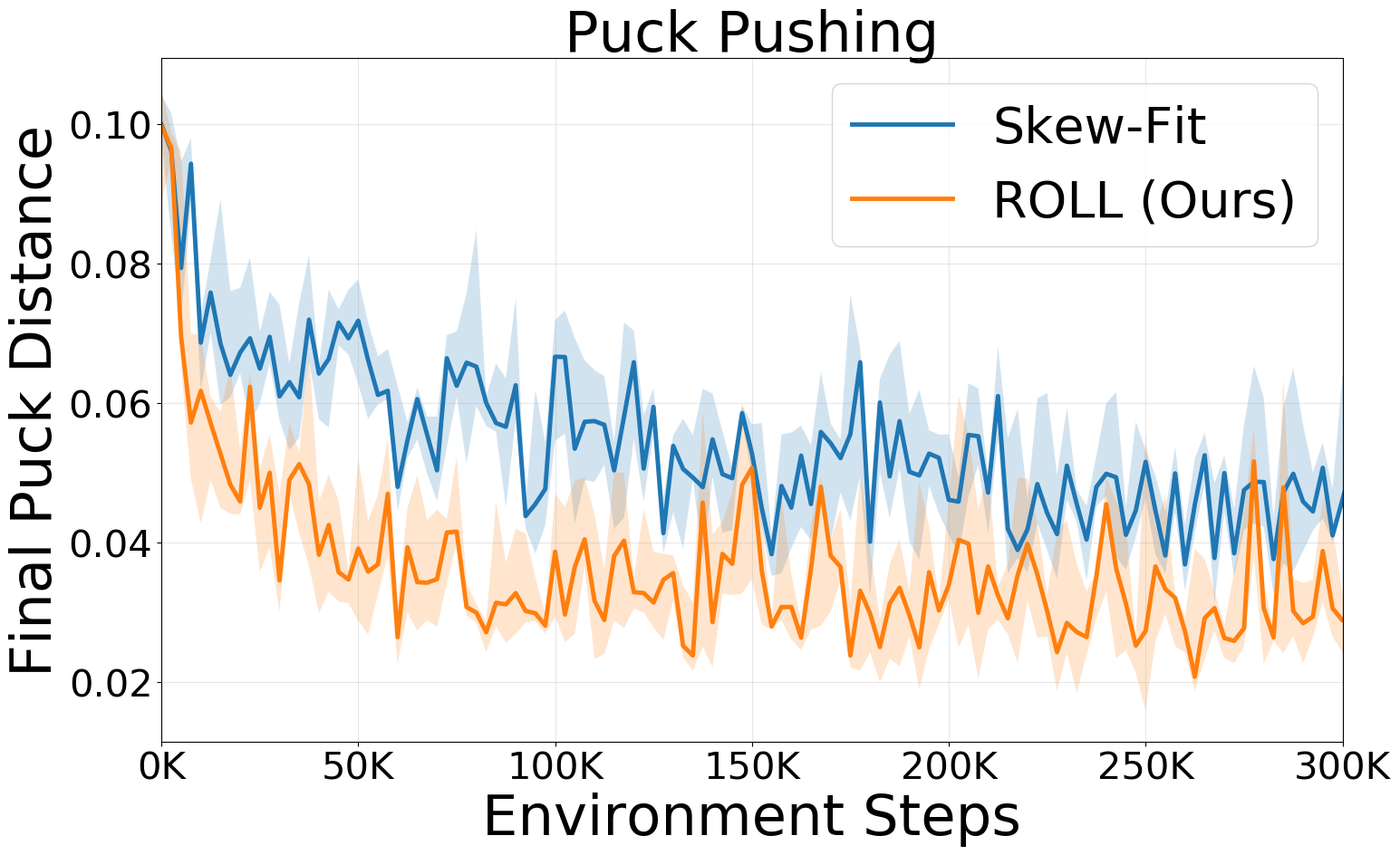}
    \includegraphics[width=0.315\textwidth]{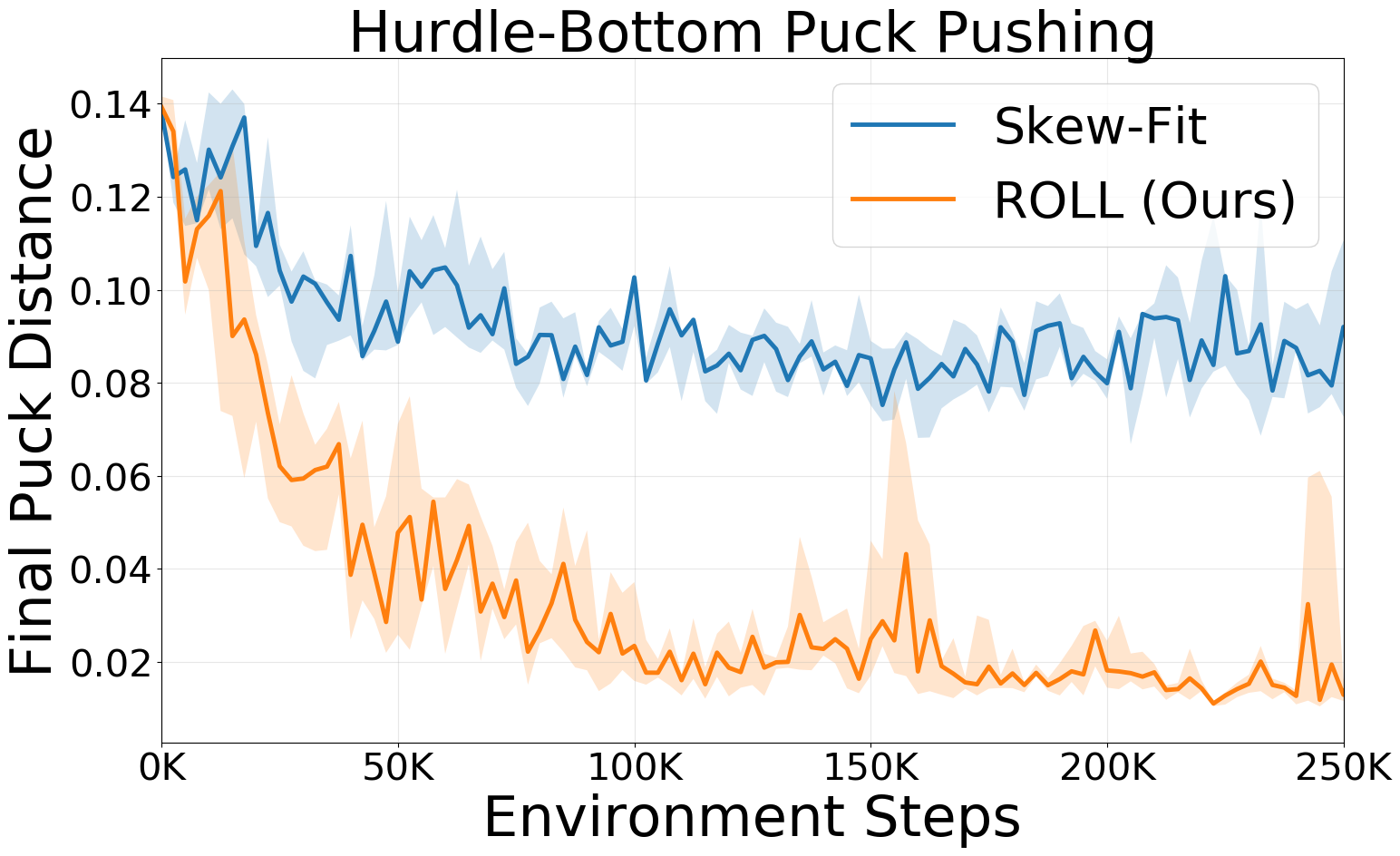}
    \includegraphics[width=0.315\textwidth]{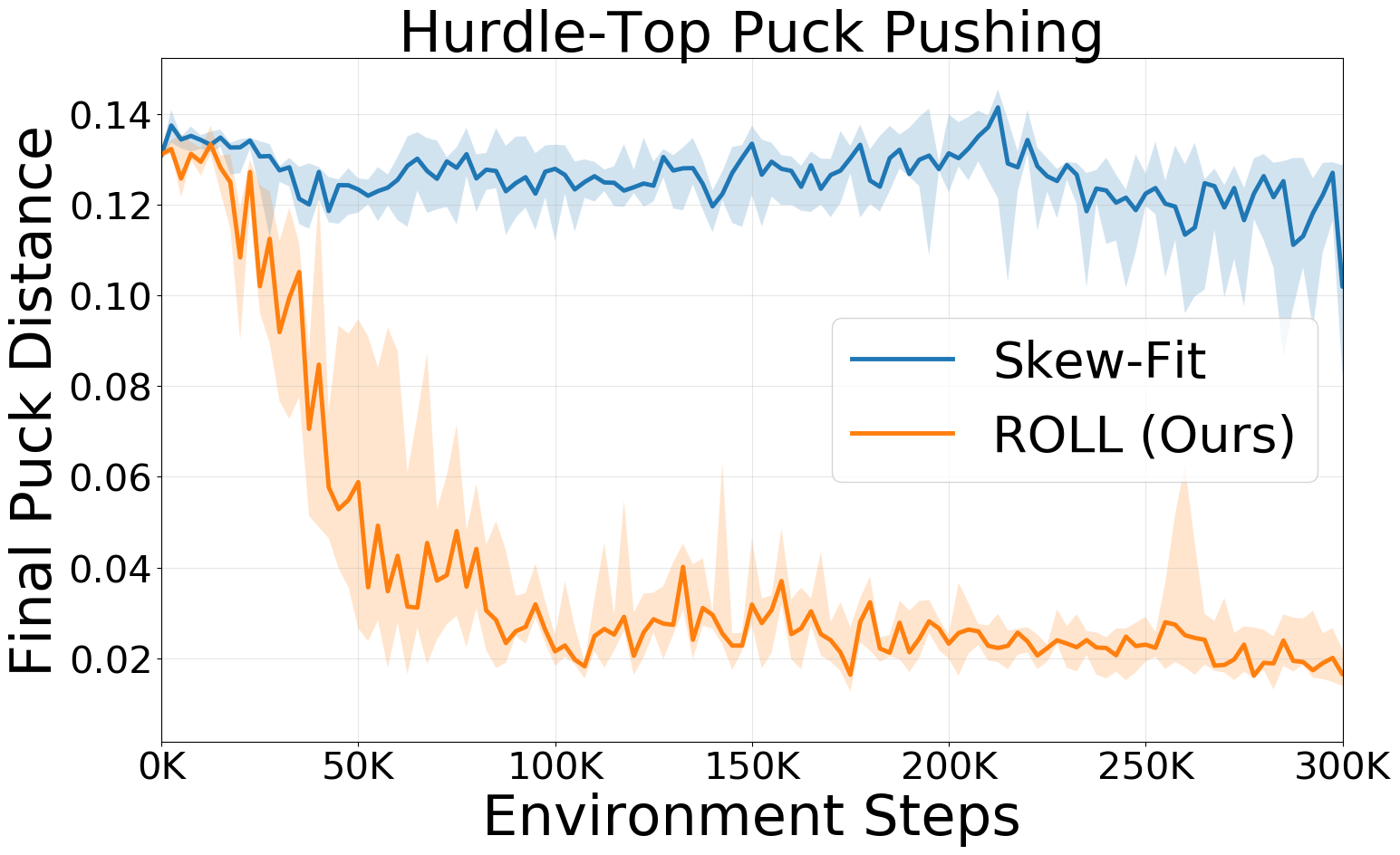}
    \includegraphics[width=0.315\textwidth]{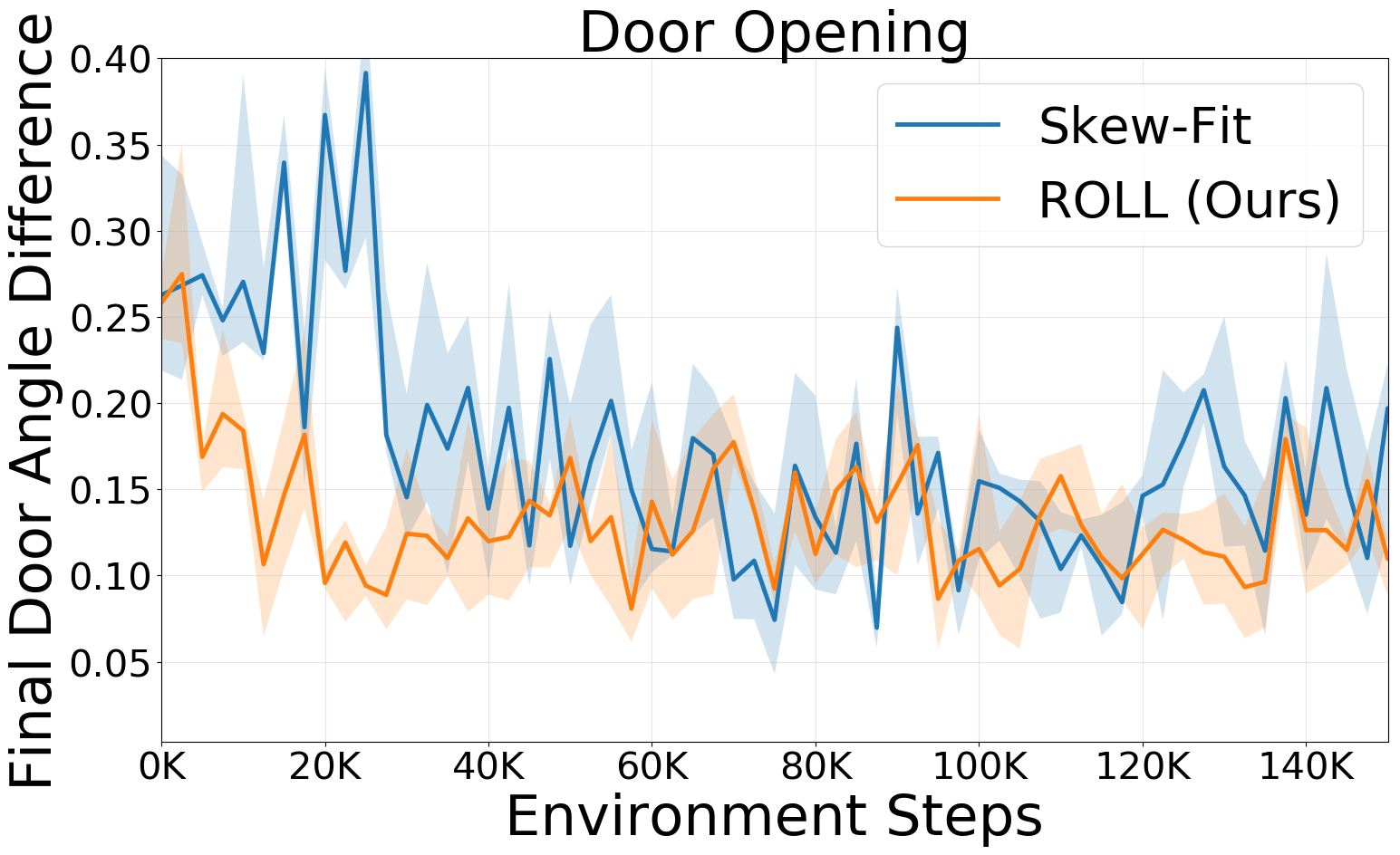}
    \includegraphics[width=0.315\textwidth]{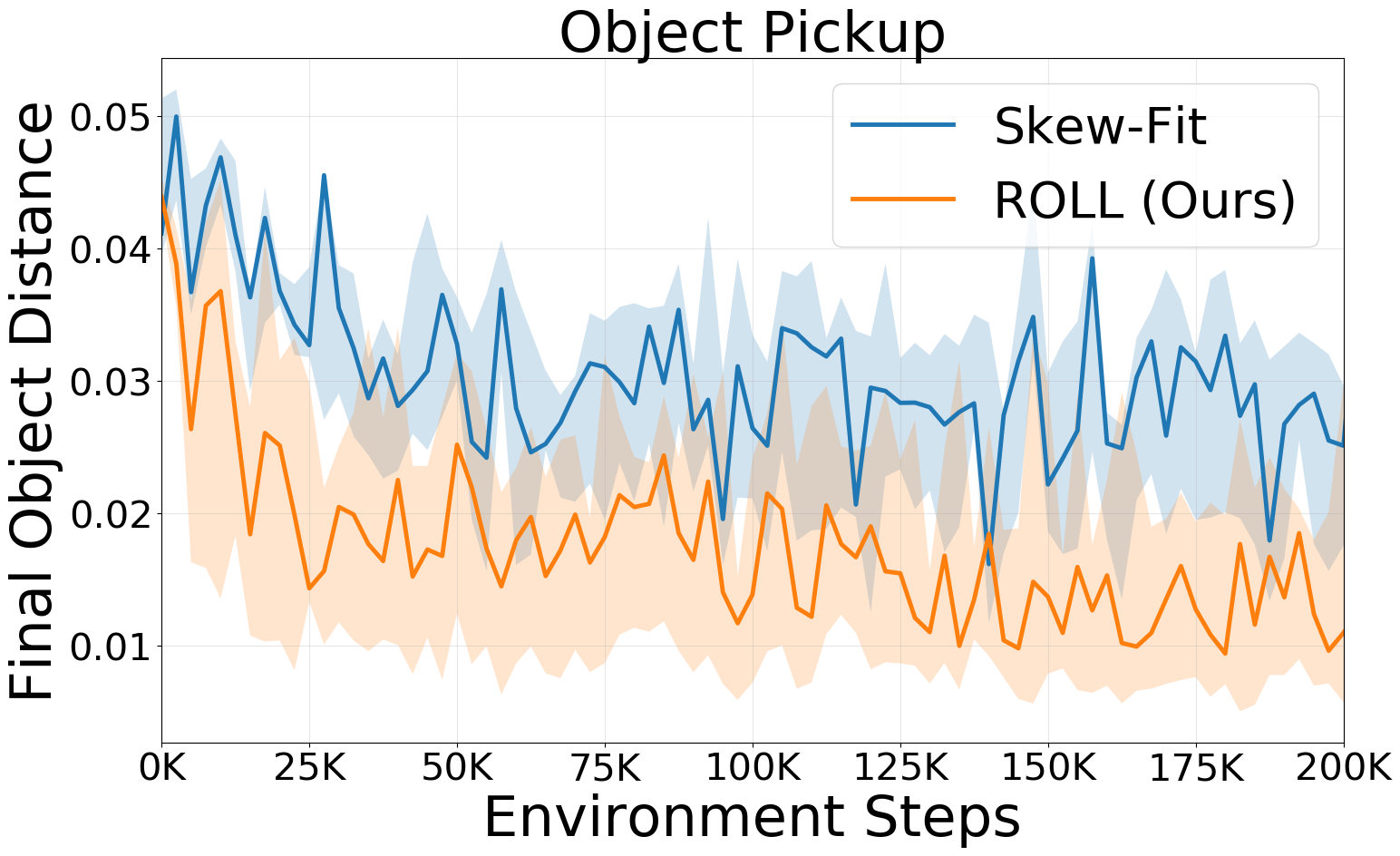} 
    \caption{Learning curves on 5 simulation tasks. The solid line shows the median of 6 seeds, and the dashed region shows the 25\% and  75\% percentile.}
    \label{fig:learning_curves}
\end{figure}

\begin{figure}
    \centering
    \includegraphics[width=0.26\textwidth]{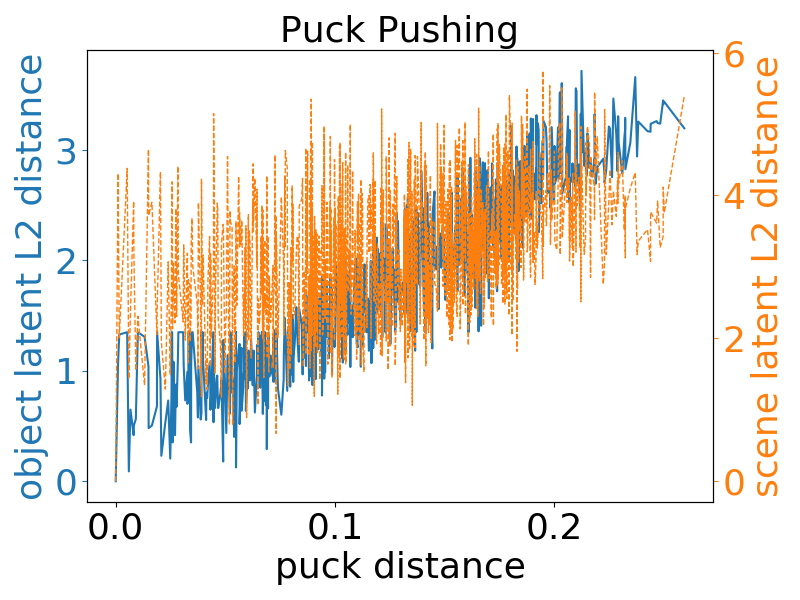}
    \includegraphics[width=0.26\textwidth]{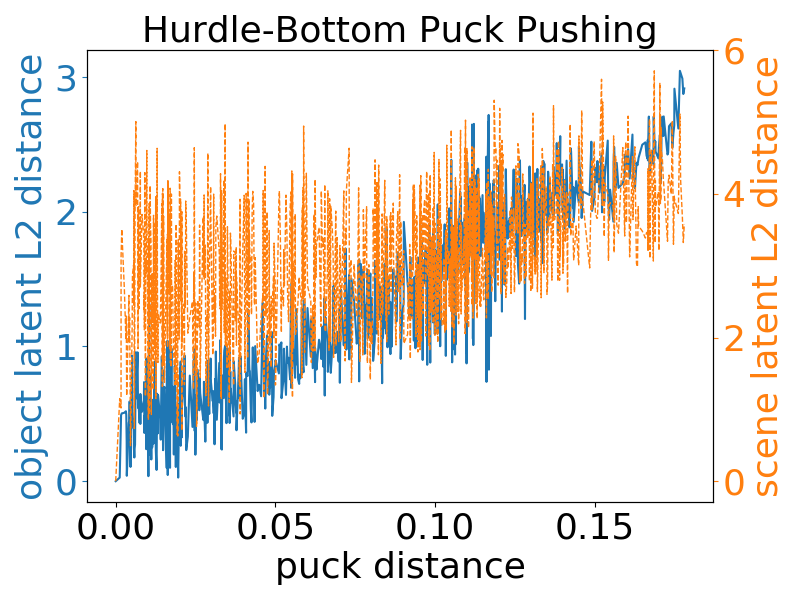}
    \includegraphics[width=0.26\textwidth]{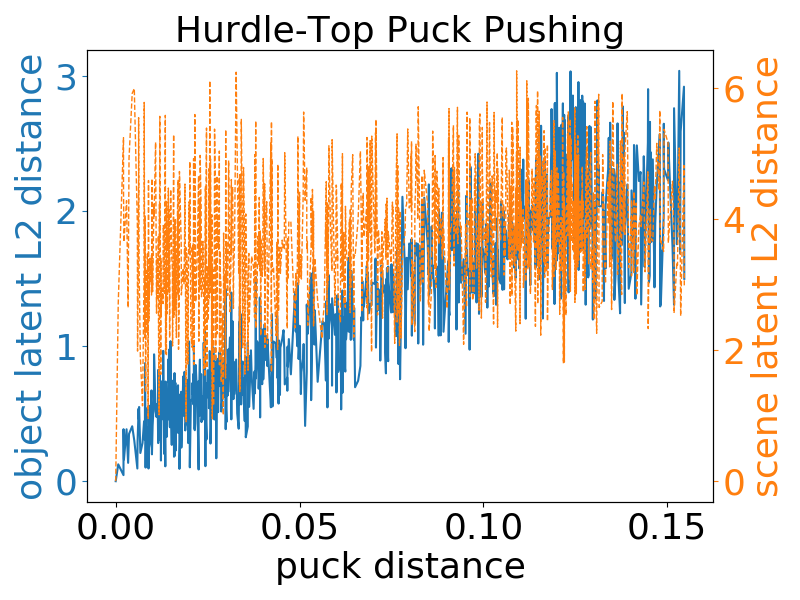}
    \includegraphics[width=0.26\textwidth]{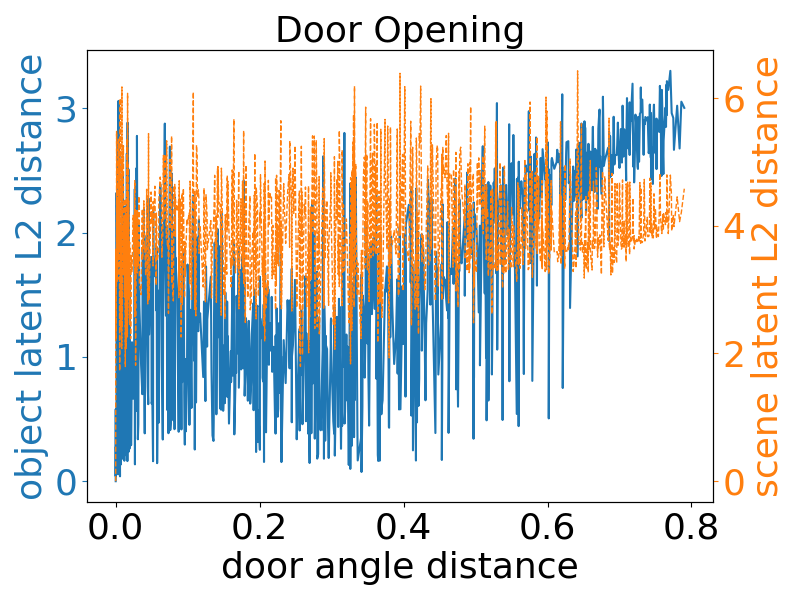}
    \includegraphics[width=0.26\textwidth]{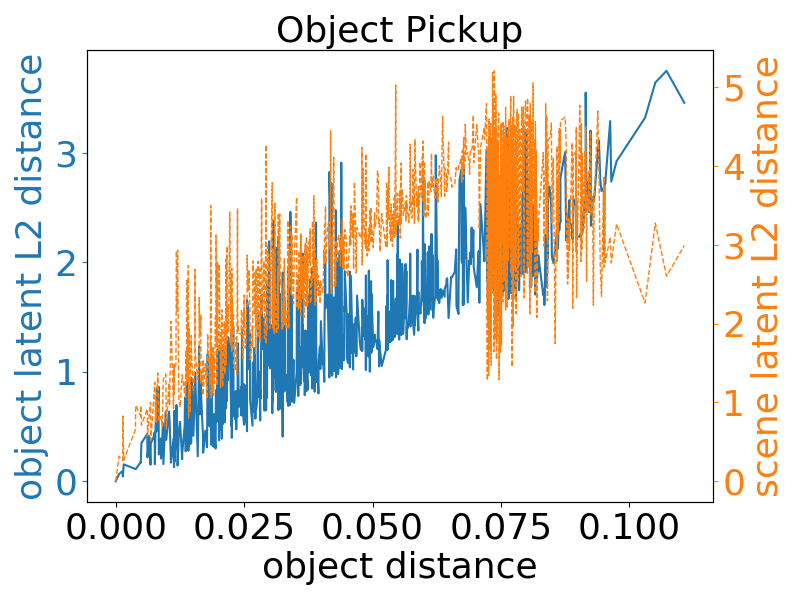}
    \caption{Comparison of the manipulated object distance (x-axis) against the distance in scene latent embedding (orange) and the distance in object latent embedding (blue). As can be seen, the object latent distance  correlates much more strongly with the manipulated object distance than the scene-VAE latent distance used in prior work.
    }
    \label{fig:better_reward}
\end{figure}

\begin{figure}[t]
    \centering
    \includegraphics[width=0.43\textwidth]{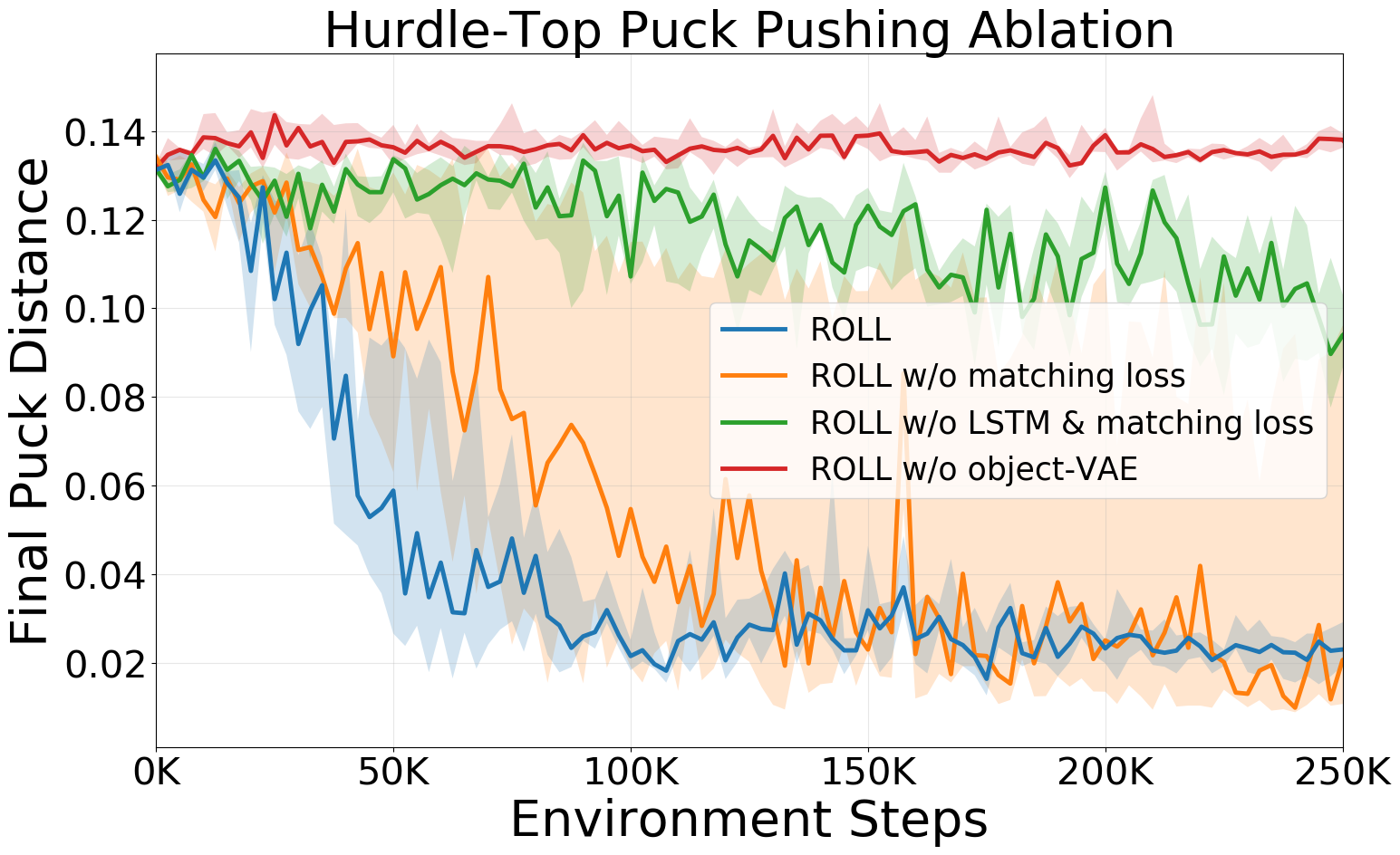} 
    \includegraphics[width=0.23\textwidth]{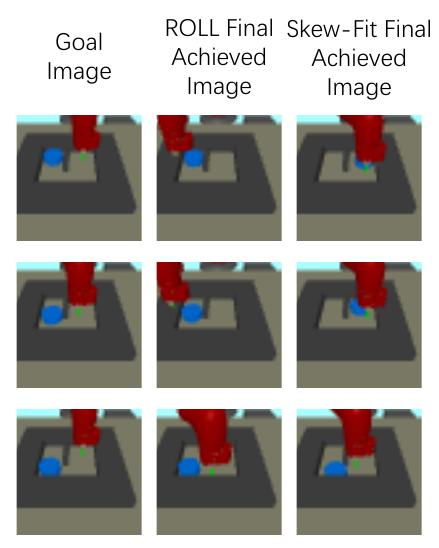} 
   
    \caption{Left: Learning curves on ablations of not using the object-VAE, the LSTM, or the matching loss on the Hurdle-Top Puck Pushing task. Right: 
    Policy visualizations in the Hurdle-Top Puck Pushing task. We see that Skew-Fit often ignores the puck and only chooses to align the arm, while \algo always successfully aligns the puck.
    } 
    \label{fig:occlusion_analysis}
\end{figure}

\textbf{\algo significantly outperforms the baseline.} Figure~\ref{fig:learning_curves} shows the learning curves of all tasks. The evaluation metric is the object distance (e.g., puck distance or door angle distance) between the object in the goal image and the final object state achieved by the policy. We can see that \algo outperforms the Skew-Fit baseline by a  large margin in all tasks except for Door Opening, where it achieves a similar final performance. For all other tasks, \algo not only obtains better final performance but also learns dramatically faster. We note that Skew-Fit performs especially poorly in the Hurdle Puck Pushing tasks. This is because, when provided a goal image where the arm and puck are at different locations, the policy learned by Skew-Fit always aligns the arm instead of the puck (see Figure~\ref{fig:occlusion_analysis}, right). Instead, by focusing on the object in the reward computation and goal conditioning, the policy learned by \algo always aligns the puck and ignores the arm.

\textbf{Segmented images provide a better reward function.} To analyze why our method performs so well, we verify if the reward function derived from the latent space of the object-VAE is better than that derived from the scene-VAE. For a better reward function, the distance in the latent space should better approximate the distance to the real object, e.g., the puck/ball distance in the pushing/pickup tasks and the door angle distance in the door opening task. In Figure~\ref{fig:better_reward}, we plot the object distance along the x-axis and the latent distance along the y-axis, where the  distance is measured between a set of observation images and a single goal image. A good latent distance should correlate with the real object distance. As can be seen, the latent distance from the object-VAE is much more accurate and stable in approximating the real object distance in all five tasks. This verifies our intuition that the reward latent encoding should only include information from the target object, rather than the robot arm or other static parts in the scene.

\textbf{Ablation study.} To test whether each component of our method is necessary, we perform ablations of our method in the Hurdle-Top Puck Pushing task, which has large occlusions on the optimal path. We test three variants of our method: \textit{\algo without matching loss}, which does not add the matching loss to the LSTM output $z^s$ or the object-VAE latent embedding $\hat{z}^s$; \textit{\algo without LSTM and matching loss}, which does not add an LSTM after the object-VAE and uses no matching loss; \textit{\algo without object-VAE}, which replaces the object-VAE in ROLL with the scene-VAE but still uses an LSTM and the matching loss. The results are shown in the left subplot of Figure~\ref{fig:occlusion_analysis}. We can see that without the the matching loss, the policy learns slower and has very large variance; without the LSTM, the policy learns very poorly; without the object-VAE, the policy cannot learn at all. This verifies the necessity of using the object-VAE, the LSTM, and the matching loss in ROLL.

%% file: inputs/6_conclusion.tex
\section{Conclusion}
In this paper, we propose ROLL, a goal-conditioned visual self-supervised reinforcement learning algorithm that incorporates object reasoning. We segment out the objects in the scene with unknown object segmentation to train an object-VAE; this provides a better reward function and goal sampling for self-supervised learning. We further employ an LSTM and a novel matching loss to make the method robust to object occlusions. Our experimental results show that \algo significantly outperforms the baselines in terms of both final performance and learning speed. This showcases the importance of an object-centric view for learning robotic manipulation tasks. We hope our work can raise more research interest in exploring visual learning with object reasoning.

%% file: inputs/appendix.tex
\clearpage
\newpage
\appendix

\newcommand{\tabincell}[2]{\begin{tabular}{@{}#1@{}}#2\end{tabular}}
\newcommand{\theHalgorithm}{\arabic{algorithm}}
\newtheorem{prop}{Proposition}
\renewcommand{\figurename}{Supplementary Figure}
\renewcommand{\tablename}{Supplementary Table}
\newcommand{\todo}[1]{\textcolor{blue}{[\textbf{TODO:} #1]}}
\newcommand{\randint}[2]{\text{randint}(#1 #2)}
\newcommand{\unif}[2]{\text{Unif}[#1 #2]}

\section{Occlusion analysis}
\label{app:occlusion-analysis}
Here we perform more analysis on how the LSTM and matching loss enable ROLL to be robust to object occlusions. 
To analyze this, we generate a large number of puck moving trajectories in the Hurdle-Top Puck Pushing environment. Next, we train three different models on these trajectories: 
\begin{itemize}
    \item LSTM with matching loss
    \item LSTM without matching loss
    \item object-VAE (with no LSTM)
\end{itemize}
For each trajectory, we add synthetic occlusions to a randomly selected frame (by removing 85\% of the pixels) and we use the three models to compute the latent encodings of the occluded frame (using the LSTM with the previous trajectory for models 1 and 2).  We then use this embedding to retrieve the nearest neighbor frame in the collection of unoccluded trajectories whose latent embedding has the closest distance to the occluded latent embedding.  This retrieved frame allows us to visualize the position that the model ``thinks" the occluded puck is located at. For models 1 and 2, the model can use the LSTM and the previous trajectory to infer the location of the puck; for model 3, it cannot.

\begin{figure}[h]
    \centering
          \includegraphics[width=0.6\textwidth]{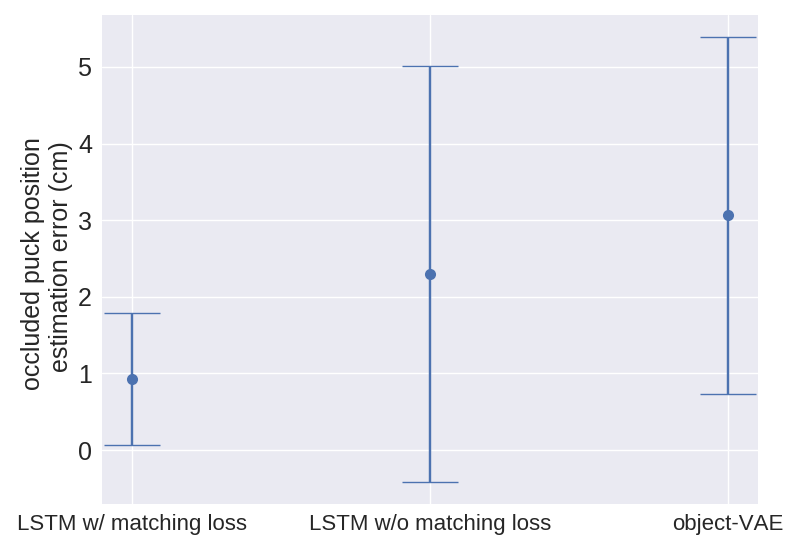}

    \caption{The estimation error of the puck position under occlusion of different methods.}
    \label{suppfig:occlusion_analysis}
\end{figure}

Finally, to evaluate this prediction, we compute the real puck distance between the location of the puck in the retrieved frame and the location of the puck in the occluded frame (using the simulator to obtain the true puck position, although this information is not available to the model).   This distance can be interpreted as the estimation error of the puck position under occlusions. We report the mean and standard deviation of the estimation errors for three models in \figurename~\ref{suppfig:occlusion_analysis}(a). As shown, using the LSTM + matching loss achieves the lowest average estimation error of roughly just 1cm, while LSTM without matching loss has a larger error of 2.2 cm and object-VAE has the largest error of 3.1 cm.

\begin{figure}[h]
    \centering
          \includegraphics[width=0.8\textwidth]{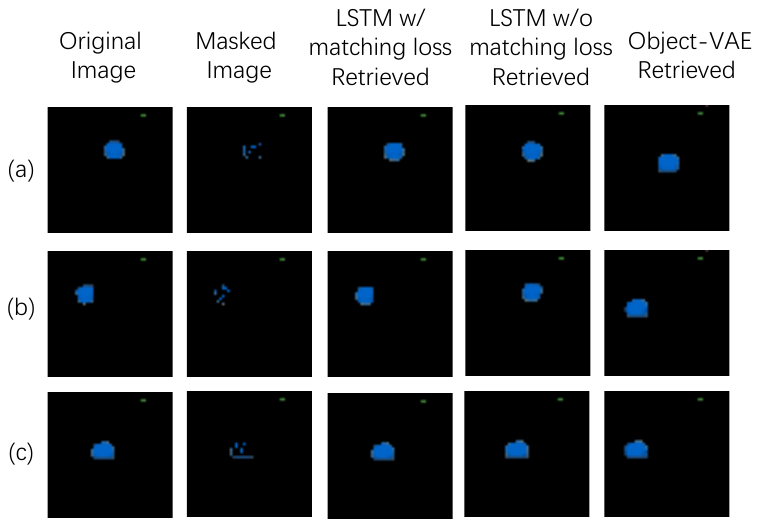}

    \caption{
    Three demonstrations on what frames different models retrieve. 
    We can see that LSTM with matching loss can accurately retrieve the true frames of the occluded puck, while LSTM without matching loss have small errors in the retrieved frames, and object-VAE has very large errors in the retrieved frames.}
    \label{suppfig:occlusion_analysis2}
\end{figure}

We can also visualize these retrievals, as shown in \figurename~\ref{suppfig:occlusion_analysis2}. We see that in all demonstrations, LSTM with matching loss almost perfectly retrieved the true unoccluded frames, while LSTM without matching loss and object-VAE retrieved incorrect frames with shifted puck positions. This shows that, even under severe occlusions (in the example, 85\% of the pixels are dropped), with the LSTM and matching loss, ROLL can still correctly reason about the location of the object.

\begin{figure}[t]
    \centering
    \begin{tabular}{cccc}
    \includegraphics[width=0.24\textwidth]{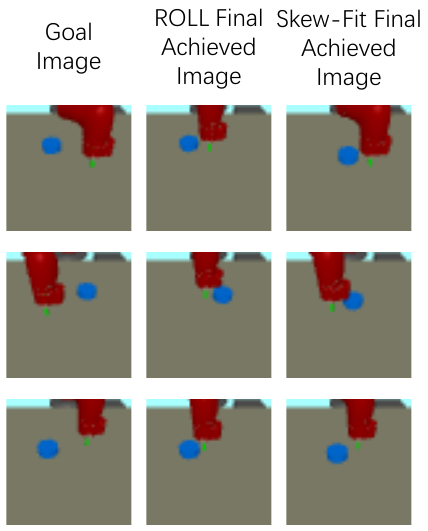}     & 
    \includegraphics[width=0.24\textwidth]{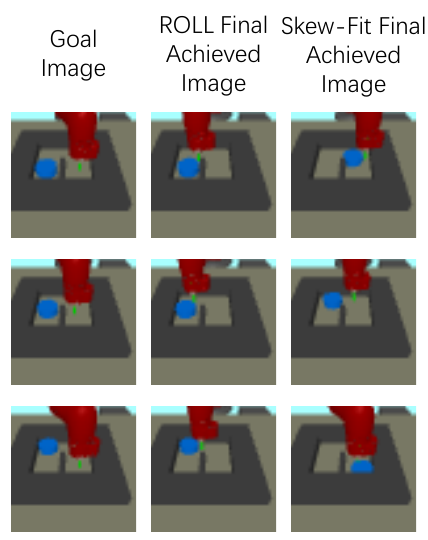}  & 
    \includegraphics[width=0.24\textwidth]{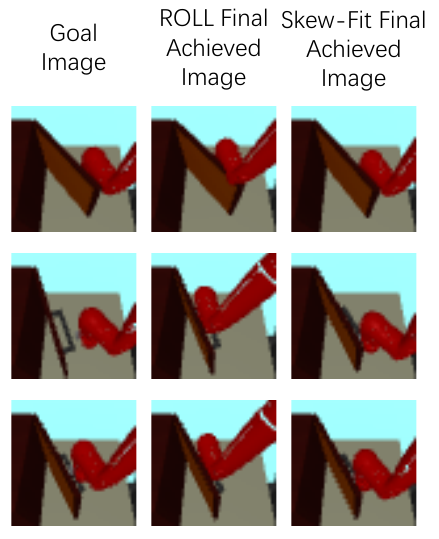} &
    \includegraphics[width=0.24\textwidth]{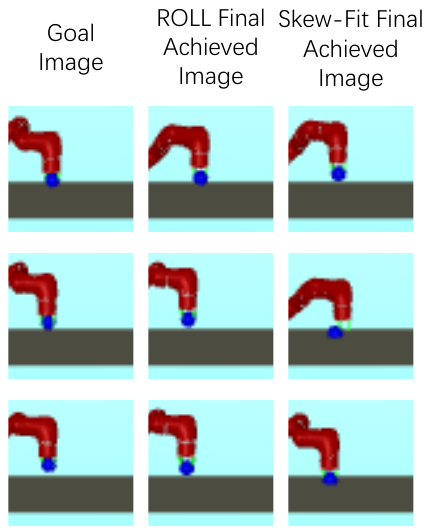} \\
    (a) & (b) & (c) & (d)
    \end{tabular}
    \caption{Policy visualizations on more tasks. (a) Puck Pushing. (b) Hurdle-Bottom Puck Pushing. (c) Door Opening. (d) Object Pickup.}
    \label{supfig:policy_visulizations}
\end{figure}

\begin{figure}[t]
    \centering
    \begin{tabular}{cc}
    \includegraphics[width=0.45\textwidth]{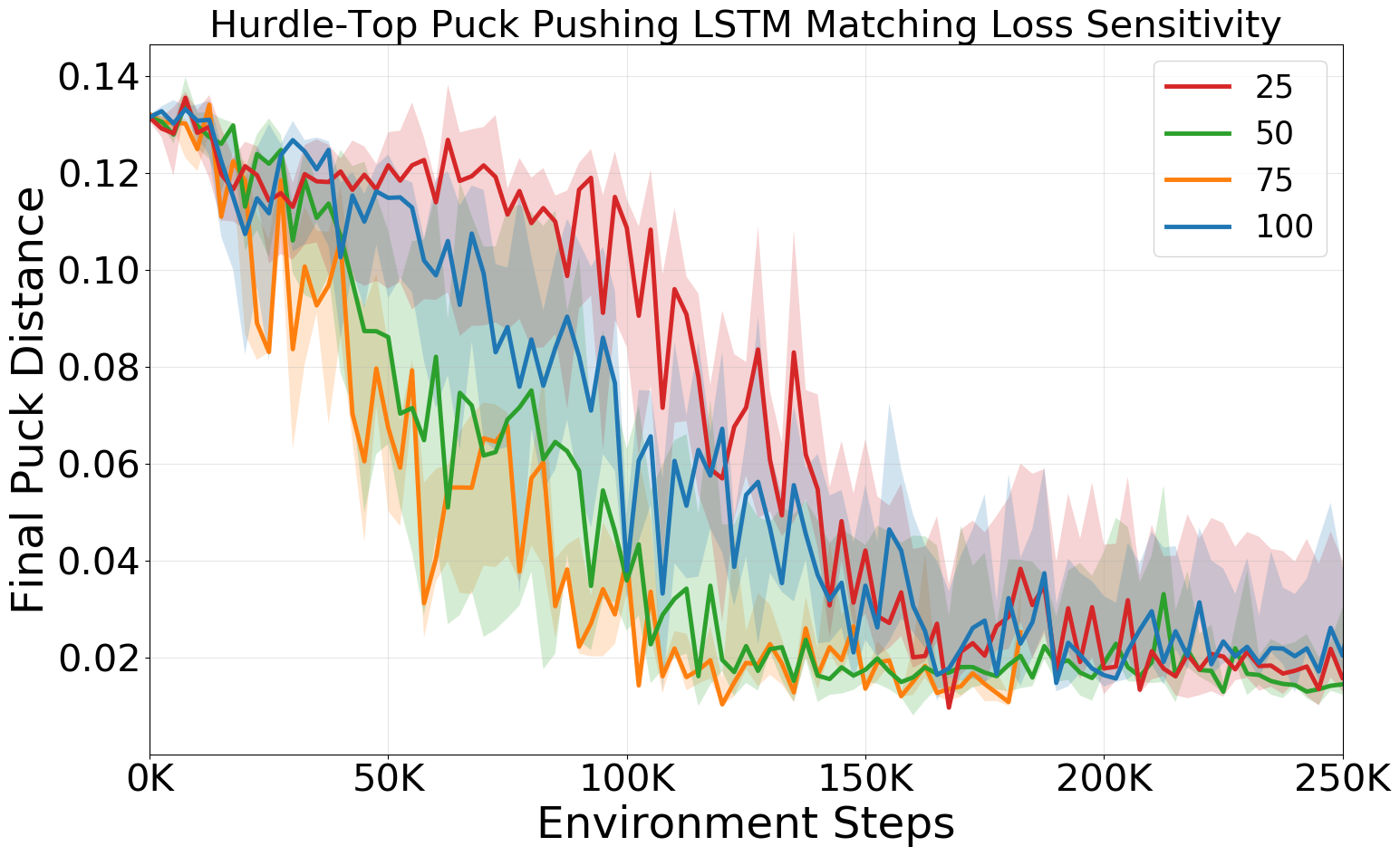}     & 
    \includegraphics[width=0.45\textwidth]{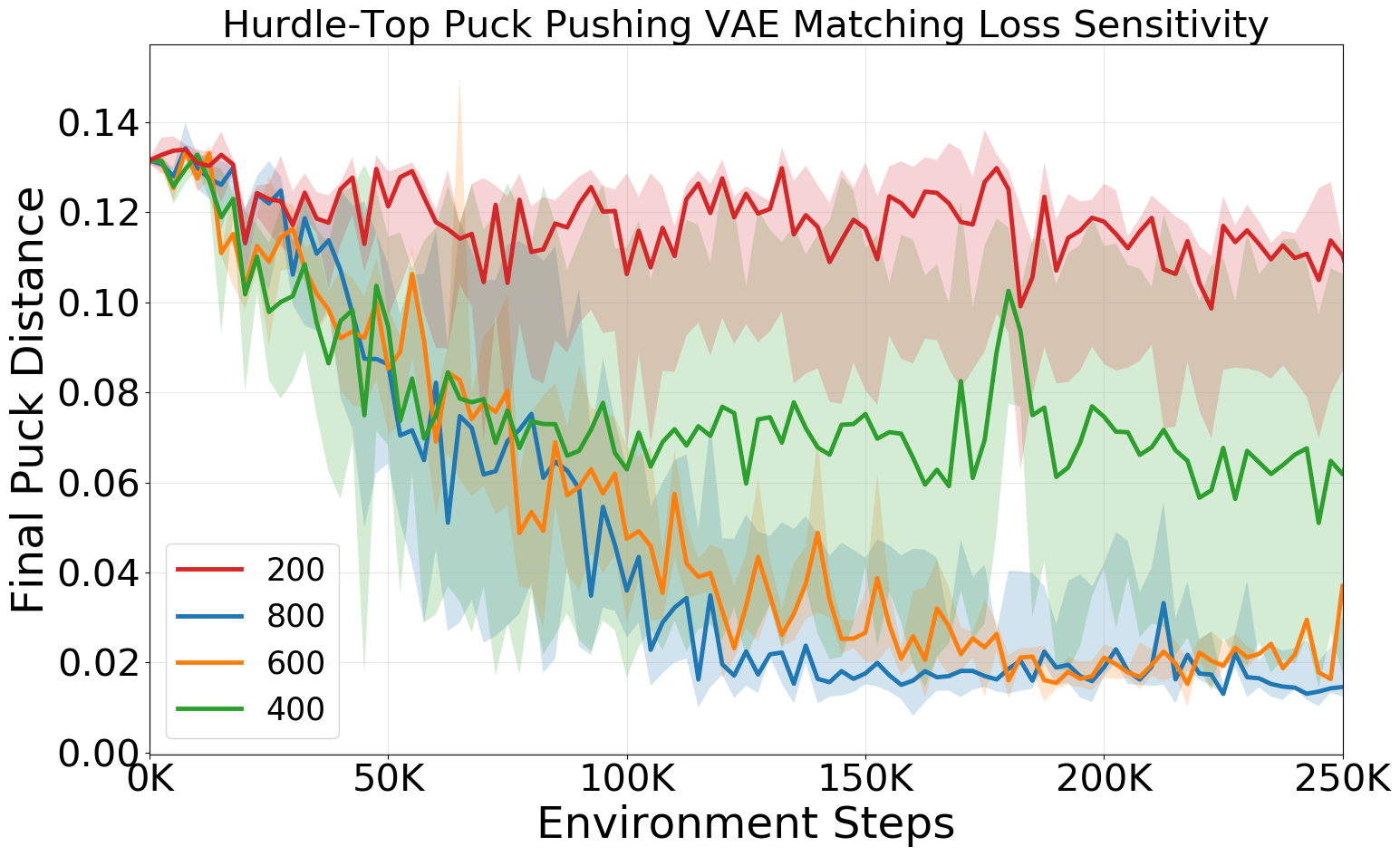}  \\ 
    (a) & (b) \\
    \includegraphics[width=0.45\textwidth]{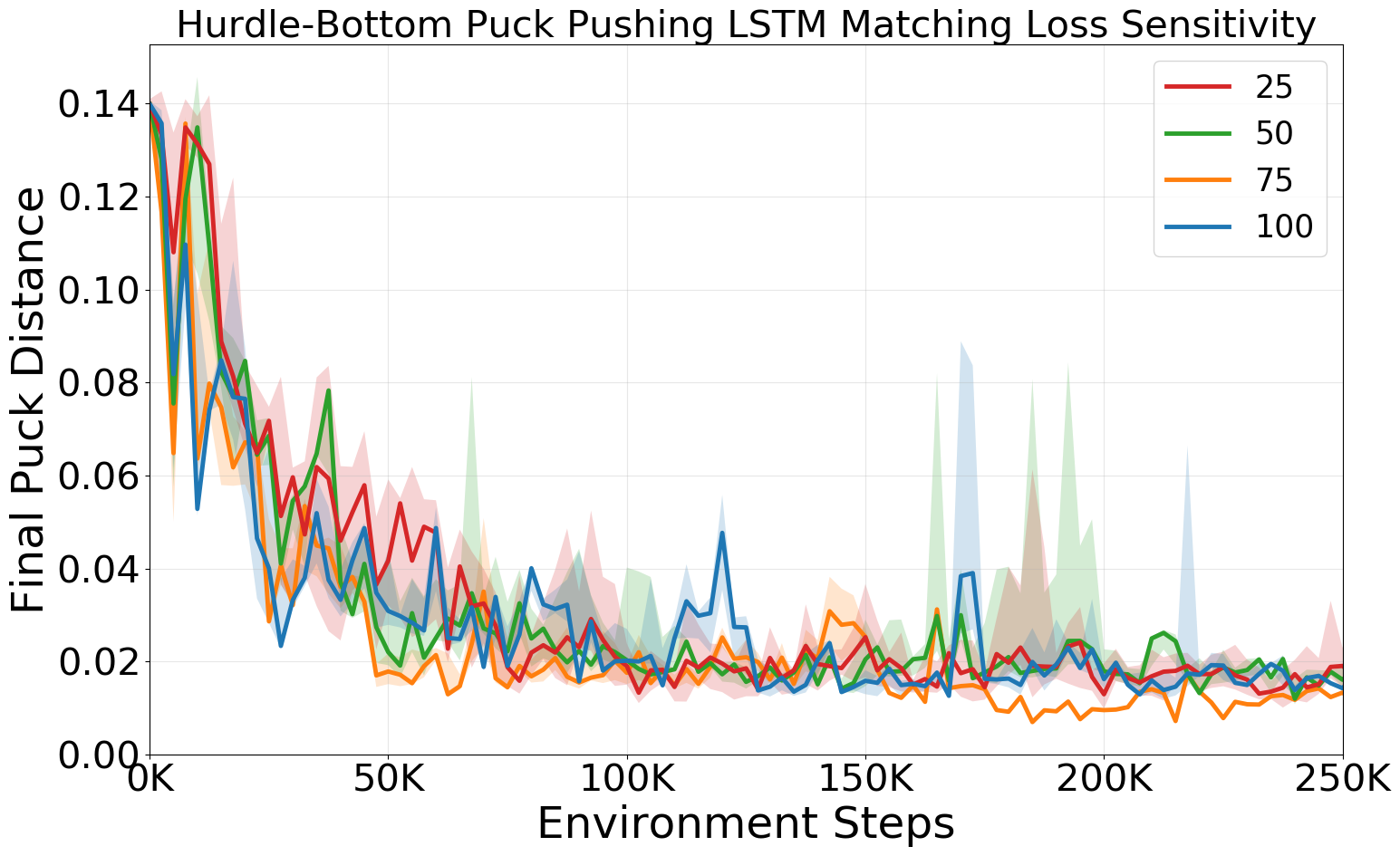}  &
    \includegraphics[width=0.45\textwidth]{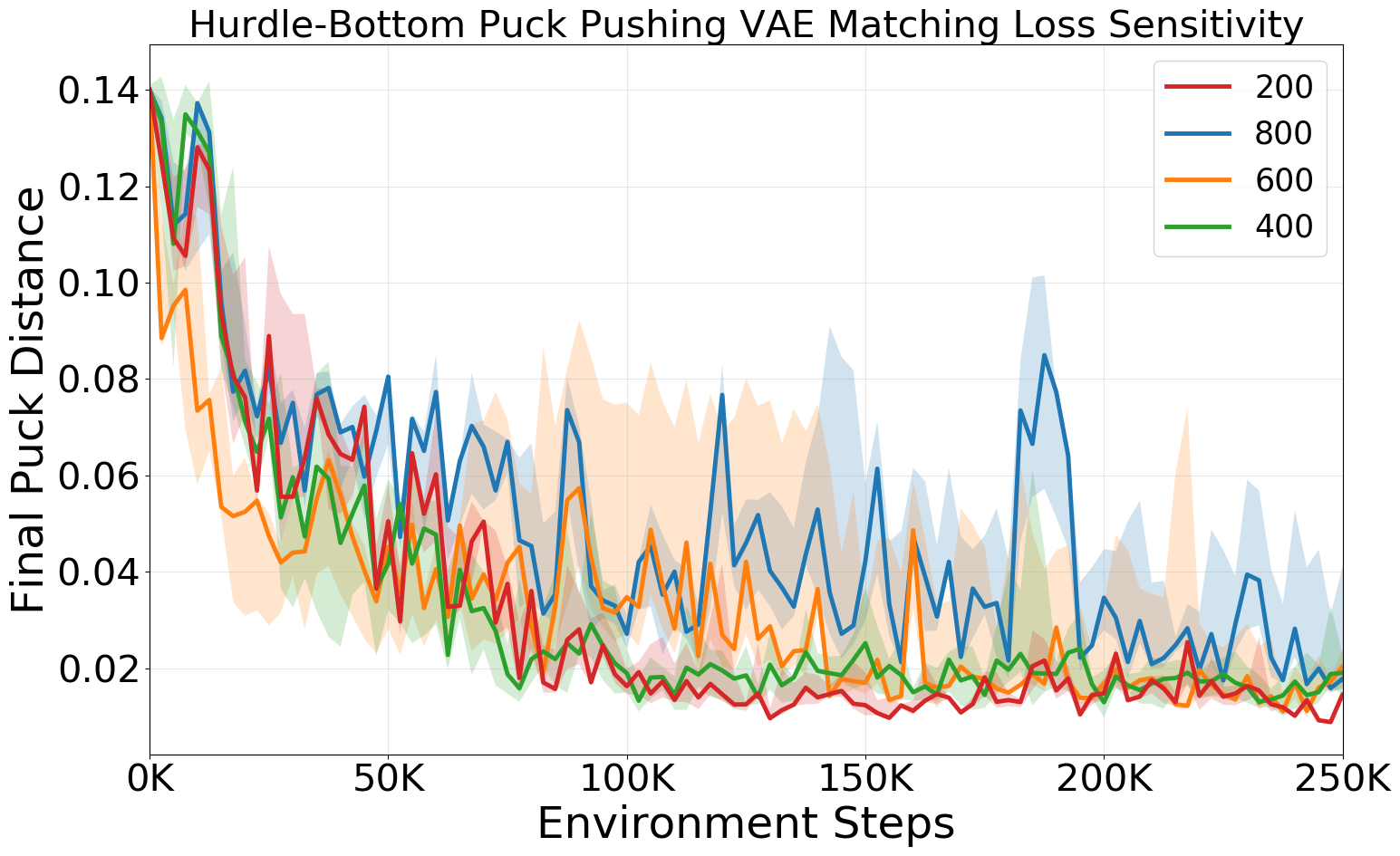}  \\
    (c) & (d)
    \end{tabular}
    \caption{Sensitivity of ROLL to the LSTM / VAE matching loss coefficient on the Hurdle-Top Puck Pushing and Hurdle-Bottom Puck Pushing tasks. (a) \& (b): results on the Hurdle-Top Puck Pushing task. (c) \& (d): results on the Hurlde-Bottom Puck Pushing task. (a) \& (c): test the sensitivity to the VAE matching loss coefficient. (b) \& (d): test the sensitivity to the LSTM matching loss coefficient. For (a), the VAE matching loss coefficient is fixed at 800; For (b), the LSTM matching loss coefficient is fixed at 50. For (c), the VAE matching loss coefficient is fixed at 400. For (d), the LSTM matching loss coefficient is fixed at 25.}
    \label{suppfig:matching_sensitivity}
\end{figure}

\section{More policy visualizations}
\label{app:more-policy-visual}
\figurename~\ref{supfig:policy_visulizations} shows more policy visualizations of ROLL and Skew-Fit. We can see that in most cases, ROLL achieves better manipulation results than Skew-Fit, aligning the object better with the target object position in the goal image.  Skew-fit does not reason about objects and instead embeds the entire scene into a latent vector; further, Skew-fit does not reason about occlusions. Videos of the learned policies on all tasks can be found on the project website.

\section{Sensitivity on matching loss coefficient}
\label{app:sensitivity-of-matching-loss}
We further test how sensitive ROLL is to the matching loss coefficient, on the Hurdle-Top Puck Pushing and the Hurdle-Bottom Puck Pushing task. The result is shown in \figurename~\ref{suppfig:matching_sensitivity}. From the results we can see that ROLL is only sensitive to the VAE matching loss coefficient when the task has large occlusions, i.e., in the Hurdle-Top Puck Pushing task. We also observe that in this task, the larger the VAE matching loss coefficient, the better the learning results. 
ROLL is more robust to the VAE matching loss coefficient in the Hurdle-Bottom Puck Pushing task, and we observe that larger VAE matching loss coefficients lead to slightly worse learning results. This is because the Hurdle-Bottom Puck Pushing task has a very small chance of object occlusions; thus too large of a VAE matching loss coefficient might instead slightly hurt the learned latent embedding. An intermediate VAE matching loss of 600 appears to perform well for both tasks.
Additionally, we see that
ROLL is quite robust to the LSTM matching loss coefficient in both tasks.

\section{Details on unknown object segmentation}
\label{app:detail-segmentation}
We now detail how we train the background subtraction module and the robot segmentation network in unknown object segmentation.

To obtain a background subtraction module, we cause the robot to perform random actions in an environment that is object free, and we record images during this movement. We then train a background subtraction module using the recorded images. Specifically, we use the Gaussian Mixture-based Background/Foreground Segmentation algorithm~\cite{GMM_bg_subtraction1, GMM_bg_subtraction2} implemented in OpenCV~\cite{opencv_library}. In more detail, we use \texttt{BackgroundSubtractorMOG2} implemented in OpenCV. We record 2000 images of the robot randomly moving in the scene, set the tracking history of \texttt{BackgroundSubtractorMOG2} to 2000, and then train it on these images with an automatically chosen learning rate and variance threshold implemented by OpenCV. The background subtraction module is fixed after this training procedure.

\begin{figure}[h]
    \centering
    \begin{tabular}{ccc}
    \includegraphics[width=0.15\textwidth]{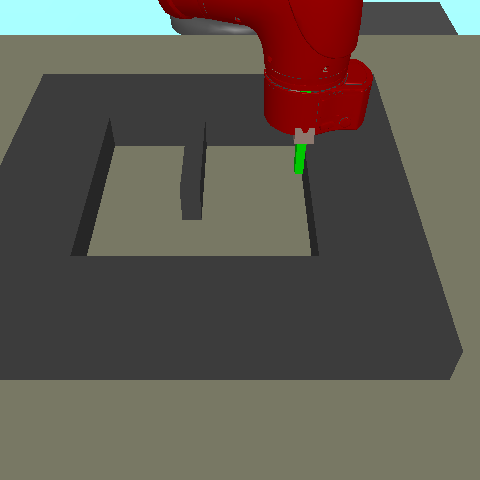} &
    \includegraphics[width=0.15\textwidth]{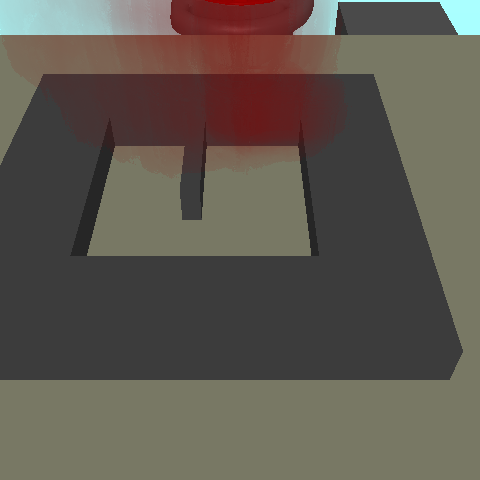} &
    \includegraphics[width=0.15\textwidth]{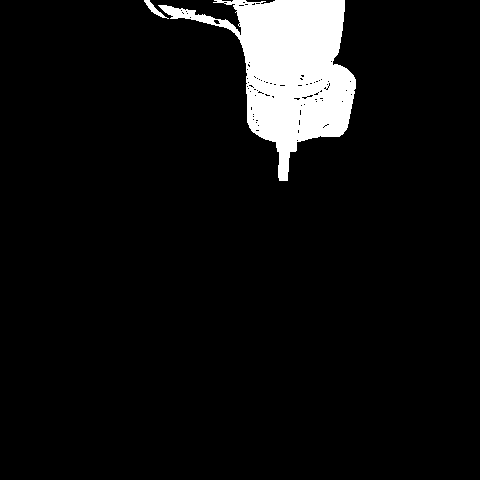} \\
    (a)  & (b) & (c)
    \end{tabular}
    \caption{(a) RGB input (b) RGB background model learned by \texttt{BackgroundSubtractorMOG2} (c) Predicted foreground on the Hurdle-Top Puck Pushing task.}
    \label{supfig:rgb_bg_model}
\end{figure}

The \texttt{BackgroundSubtractorMOG2} learns to classify non-moving objects in a scene as background, and any pixel values that fall outside a variance threshold of the Gaussian Mixture Model are classified as foreground. Illustrations of the learned background model are shown in \figurename~\ref{supfig:rgb_bg_model}. At test time, objects placed in the environment appear as foreground as their pixel values fall outside the threshold.  Similarly, the robot also appears as foreground, for the same reason.  We address this issue using a robot segmentation network, explained in detail below. 

In order to remove the robot from the scene, we train a robot segmentation network.
To generate training labels, we use the trained background subtraction module, described above.  Using the same dataset used to train the OpenCV background subtraction module (with no objects in the scene), we run the OpenCV background subtraction module.  Points that are classified as foreground belong to the robot.  We use this output as ``ground-truth" segmentation labels.  Using these labels, we train a network to segment the robot from the background. We use U-Net~\cite{ronneberger2015u} as the segmentation network. The U-Net model we use has 4 blocks of down-sampling convolutions and then 4 blocks of up-sampling covolutions. Every block has a max-pool layer, two convolutional layers each followed by a batch normalization layer and a ReLU activation. Each up-sampling layer has input channels concatenated from the outputs of its down-sampling counterpart. These additional features concatenated from the input convolutions help propagate context information to the higher resolution up-sampling layers. The kernel size is 3x3, with stride 1 and padding 1 for all the convolutional layers. 

We train the network using a  binary cross entropy loss. The optimization is performed using Nesterov momentum gradient descent for 30 epochs with a learning rate of 1e-3, momentum of 0.9, and a weight decay of 5e-4. 

One potential issue of the above method is that the robot segmentation module has only been trained on images without objects in the scene.  We find that adding synthetic distractors to the scene helps to improve performance.  In this work, we use distractors created by masks of objects similar to those at test time.  In future work, we will instead use diverse distractors taken from the COCO dataset~\cite{lin2014microsoft}.

\section{Simulated task details}
\label{app:task-detail}
\begin{figure}[h]
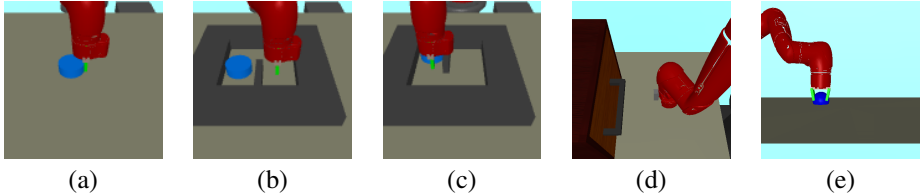

    \centering
    \begin{tabular}{ccccc}
        \includegraphics[width=0.15\textwidth]{images/env_visual/SawyerPushNIPSEasy-v0.png} &
        \includegraphics[width=0.15\textwidth]{images/env_visual/SawyerPushHurdleMiddle-v0.png}& 
        \includegraphics[width=0.15\textwidth]{images/main/occlusion_hurdle.png} &
        \includegraphics[width=0.15\textwidth]{images/env_visual/door_env_1.png}&
        \includegraphics[width=0.15\textwidth]{images/env_visual/pickup_env_2.png} \\
        (a)  & (b) & (c)  &(d) & (e) 
    \end{tabular}
    \caption{The robot view of different tasks: (a) Puck Pushing (b) Hurdle-Bottom Puck Pushing (c) Hurdle-Top Puck Pushing (d) Door Opening (e) Object Pickup. 
    }
    \label{supfig:environment_imgs}
\end{figure}

All the tasks are simulated using the MuJoCo~\cite{todorov2012mujoco} physics engine. The Puck Pushing, Door Opening, and Object Pickup tasks are identical to those used in Skew-Fit~\cite{skewfit}. Illustrations of the environments are shown in \figurename~\ref{supfig:environment_imgs}(a), (d), and (e). We also added two additional environments with obstacles and challenging occlusions, shown in \figurename~\ref{supfig:environment_imgs}(b) and (c).

For the coordinates used in the puck pushing tasks, the x-axis goes towards the right direction, and the y-axis goes towards the bottom direction in \figurename~\ref{supfig:environment_imgs}.

\noindent\textbf{Puck Pushing:} A 7-DoF
Sawyer arm must push a small puck on a table to various target positions. The agent controls the arm by
commanding the $\delta x, \delta y$ movement of the end effector (EE). The
underlying state is the EE position $e$ and the puck position $p$.
The evaluation metric is the distance between the goal and
final achieved puck positions. The hand goal/state space is a box $[-0.1, 0.1] \times [0.55, 0.65]$. The puck goal/state space is a box $[-0.15, 0.15] \times [0.5, 0.7]$.
The action space ranges in the interval $[-1, 1]$ in the $x, y$ dimensions. The arm is always reset to $(-0.02, 0.5)$ and the puck is always reset to $(0, 0.6)$.

\noindent\textbf{Hurdle-Bottom Puck Pushing:} The task is similar to that of Puck Pushing, except we add hurdles on the table to restrict the movement of the puck. The coordinates of the inner corners of the hurdle are top-left $(0.11, 0.52)$, top-right $(-0.025, 0.52)$, bottom-left $(0.11, 0.67)$, bottom-right $(-0.025, 0.67)$. The left corridor to the middle-hurdle has a width of $0.06$, and the right corridor to the middle-hurdle has a width of $0.065$. The corridor up to the hurdle has a width of $0.07$. The middle-hurdle has a width of $0.01$ and a length of $0.08$. The arm is always reset to the location of $(-0.02, 0.5)$ and the puck is reset to $(-0.02, 0.54)$. The puck goal space is $[0.1, 0.11] \times [0.55, 0.65]$ (i.e., roughly the range of the left corridor). The hand goal space is $[-0.025, 0] \times [0.6, 0.65]$ (i.e., roughly the bottom part of the right corridor).

\noindent\textbf{Hurdle-Top Puck Pushing: } The task is similar to that of Hurdle-Bottom Puck Pushing, except the position of the middle-hurdle is flipped. 
The arm is always reset to $(-0.02, 0.5)$ and the puck is randomly reset to be in $[-0.04, -0.02] \times [0.55, 0.63]$ (i.e., roughly the top part of the right corridor). The puck goal space is $[0.1, 0.11] \times [0.55, 0.6]$ (i.e., roughly the top part of the left corridor), and the hand goal space is $[-0.03, 0] \times [0.54, 0.6]$ (i.e., roughly the top part of the right corridor).

\noindent\textbf{Door Opening}: A 7-DoF Sawyer
arm must pull a door on a table to various
target angles. The agent control is the same as in Puck Pushing, i.e., the $\delta x, \delta y$ movement of the end effector. The
evaluation metric is the distance between the goal and final
door angle, measured in radians. In this environment, we do
not reset the position of the hand or door at the end of each
trajectory. The state/goal space is a $5\times 20 \times 15$ cm$^3$ box in
the $x, y, z$ dimension respectively for the arm and an angle
between $[0, .83]$ radians for the door. The action space ranges in the
interval $[-1, 1]$ in the $x$, $y$ and $z$ dimensions.

\noindent\textbf{Object Pickup}: A 7-DoF Sawyer
arm must pick up an object on a table to various
target positions. The object
is cube-shaped, but a larger intangible sphere is overlaid on
top so that it is easier for the agent to see. Moreover, the
robot is constrained to move in 2 dimension: it only controls
the y, z arm positions. The x position of both the arm and
the object is fixed. The evaluation metric is the distance
between the goal and final object position. For the purpose
of evaluation, 75\% of the goals have the object in the air and
25\% have the object on the ground. The state/goal space for
both the object and the arm is 10cm in the y dimension and
13cm in the z dimension. The action space ranges in the
interval $[-1, 1]$ in the y and z dimensions.

\section{Implementation details}
\label{app:implementation-detail}
Our implementation of ROLL is based on the open-source implementation of Skew-Fit in RLkit\footnote{https://github.com/vitchyr/rlkit}. For all simulated tasks, the image size is  $48 \times 48$. A summary of the task specific hyper-parameters of ROLL are shown in \tablename~\ref{supptable:different-hyperparameters}. The first 4 rows use the same hyper-parameters as in Skew-Fit~\cite{skewfit}, and the last two rows are hyper-parameters that we tune in ROLL, described in more detail below. A summary of the task-independent hyper-parameters of ROLL for training the scene-VAE, the object-VAE and the LSTM are shown in \tablename~\ref{supptable:lstm-parameters}, with detailed descriptions in the text below.

\subsection{Network Architectures}
We first describe the network architecture of each component in ROLL.  For the scene-VAE, we use the same architecture as that in Skew-Fit. In more detail, the VAE encoder has three convolutional layers with kernel
sizes: $5\times5$, $3\times3$, and $3\times3$, number of output filters: $16$,
$32$, and $64$; and strides: $3$, $2$, and $2$. The final feature map is mapped by a fully connected layer into a final feature vector of size $576$, and then we have another fully
connected layer to output the final latent embedding. The decoder has a fully connected layer that maps the latent embedding into a vector of dimension $576$.  This vector is then reshaped into a feature map of size $3 \times 3 \times 64$. The decoder has 3 de-convolution layers with kernel sizes $3 \times 3$, $3 \times 3$, $6 \times 6$, number of output filters $32$, $16$, and $3$, and strides $2$, $2$, and $3$. 

The object-VAE has almost the same architecture as the scene-VAE. However, the object-VAE has a simpler task that it only needs to encode the segmented object, rather than the entire scene.  Thus, 
we use a smaller final feature vector -- the final encoder feature vector is of size $6$ instead of $576$ as used in the scene-VAE.
Both VAEs have a Gaussian decoder
with identity variance; thus the log likelihood loss used to train the decoder is equivalent to a mean-squared error loss.

For the scenc-VAE,  we vary the value of $\beta$ as in Skew-Fit for different tasks (shown in \tablename~\ref{supptable:different-hyperparameters}). For the object-VAE, we use the same value of 20 for $\beta$, and the same latent dimension size of 6 in all tasks, as shown in \tablename~\ref{supptable:lstm-parameters}.
For the scene-VAE in Skew-Fit, we use a latent dimension size of 16 for the Door Opening task and the Object Pickup task, and use a latent dimension size of 4 for Puck Pushing tasks (which is the same as in the original Skew-Fit implementation). For the Puck Pushing with hurdle tasks, we use a latent dimension  of size 6. For ROLL, the scene-VAE latent dimension size is the same as that of Skew-Fit except for the Puck Pushing task, where we increase the latent dimension from 4 to 6 to make it the same as the object-VAE latent dimension. For this task, we also observe that using a latent dimension of $6$ for the scene-VAE performs better than a latent dimension of $4$ in ROLL. However, using a latent dimension of $6$ for the scene-VAE in Skew-Fit performs slightly worse than using a latent dimension of $4$ in this task.

The input to the LSTM is the latent vector from the object-VAE. The LSTM for all tasks has 2 layers and a hidden size of 128 units.

For the policy and Q-network used in SAC, we use exactly the same architecture as in Skew-Fit. For both networks, we use fully connected networks with two hidden layers of size 400 and 300 each, and use ReLU as the activation function.

\begin{table*}
    \centering
    \small
    \begin{tabular}{c|c|c|c|c|c}
    \hline
    \textbf{Hyper-parameter} & \textbf{\tabincell{c}{Puck\\Pushing}} & \textbf{\tabincell{c}{Hurdle-Bottom \\ Puck Pushing}} & \textbf{\tabincell{c}{Hurdle-Top \\ Puck Pushing}} & \textbf{\tabincell{c}{Door \\ Opening}} & \textbf{\tabincell{c}{Object \\ Pickup}}\\
    \hline
    Trajectory Length & $50$& $50$ & $50$ & $100$ & $50$ \\ \hline
    $\beta$ for scene-VAE & $20$ & $20$ & $20$ & $20$ & $30$ \\ \hline
    \tabincell{c}{Scene-VAE Latent Dimension \\ Size in Skew-Fit} & $4$ & $6$ & $6$ & $16$ & $16$ \\ \hline
    Skew-Fit $\alpha$ for scene-VAE & $-1$ & $-1$ & $-1$ & $-0.5$ & $-1$ \\ \hline
    \tabincell{c}{Scene-VAE Latent Dimension \\Size in ROLL} & $6$ & $6$ & $6$ & $16$ & $16$ \\ \hline
    VAE matching loss coefficient & $400$ & $400$ & $800$ & $50$ & $50$ \\ \hline
    \end{tabular}
\caption{Task specific hyper-parameters. The first four rows use the same hyper-parameters as in Skew-Fit. The fifth row shows that for ROLL, we increase the scene-VAE latent dimension size in Puck Pushing from 4 to 6, as to keep it the same as the object-VAE latent size that we use, which is 6 for all tasks. We also observe that using a latent dimension of $6$ for the scene-VAE performs better than using a latent dimension of $4$ in ROLL for this task. However, using a latent dimension of $6$ for the scene-VAE in Skew-Fit performs slightly worse than using a latent dimension of $4$ in this task.  The last row is one new hyper-parameter that we introduce in ROLL.}
\label{supptable:different-hyperparameters}
\end{table*}

\begin{table*}
    \centering
    \begin{tabular}{c|c}
    \hline
    \textbf{Hyper-parameter} & \textbf{Value}\\
    \hline
    Scene-VAE Batch Size (for both Skew-Fit and ROLL) & $64$ \\
    Object-VAE Batch Size & $128$ \\
    $\beta$ for Object-VAE & $20$ \\
    Obect-VAE Latent Dimension Size & $6$ \\
    LSTM Matching Loss Coefficient & $50$ \\
    \hline
    \end{tabular}
\caption{Scene-VAE, object-VAE and LSTM training hyper-parameters for all tasks.}
\label{supptable:lstm-parameters}
\end{table*}

\subsection{Training schedules}
We train the scene-VAE using the regular $\beta$-VAE loss, i.e., the image reconstruction loss and the KL regularization loss. We pre-train it using $2000$ images obtained by running a random policy for $2000$ epochs.  In each epoch we train for 25 batches with a batch size of 64 and a learning rate of $1e-3$.
We also continue to train the scene-VAE alongside during RL training, using images stored in the replay buffer. We sample images from the replay buffer using a skewed distribution as implemented in Skew-Fit. For different tasks we use different skewness $\alpha$ as shown in \tablename~\ref{supptable:different-hyperparameters}, which is the same as in Skew-Fit. For online training of the scene-VAE, Skew-Fit use three different training schedules for different tasks, and we follow the same training schedule as in Skew-Fit. For details on the training schedule, please refer to appendix C.5 of the Skew-Fit paper.

We train the object-VAE using the image reconstruction loss, the KL regularization loss, and the matching loss. For different tasks, we use different coefficients for the VAE-matching loss, as shown in \tablename~\ref{supptable:different-hyperparameters}. We pre-train the object-VAE with 2000 segmented images obtained by randomly putting the object in the scene.  The object-VAE is trained for 2000 epochs, where in each epoch we train for 25 batches with a batch size of 128.  We use a learning rate of $1e-3$. After the pre-training, the object-VAE is fixed during RL learning. 
For the synthetic occlusions we add for computing the matching loss, we randomly drop 50\% pixels in the segmented objects.

We train the LSTM using an auto-encoder loss and matching loss. For all tasks, we use the same LSTM-matching loss coefficient of $50$, as shown in \tablename~\ref{supptable:lstm-parameters}. We pre-train the LSTM on the same dataset we use to pre-train the object-VAE, using the auto-encoder loss for 2000 epochs. In each epoch we train for 25 batches with a batch size of 128 and a learning rate of $1e-3$. We continue training the LSTM during the RL learning process. During online training, the training trajectories are sampled uniformly from the SAC replay buffer, and we use both the matching loss and the auto-encoder loss to train the LSTM. We use a learning rate of $1e-3$ for training the LSTM. 
The online training scheme for LSTM is: for the first 5k time steps, we train the LSTM every 500 time steps for 80 batches, where each batch has 25 trajectories. For 5k - 50k time steps, we train the LSTM every 500 time steps for 20 batches. After 50k time steps, we train the LSTM every 1000 time steps for 20 batches.
For the SAC training schedule, we use the default values as in Skew-Fit; these values are summarized in \tablename~\ref{supptable:sac-parameters}.

\begin{table*}
    \centering
    \begin{tabular}{c|c}
    \hline
    \textbf{Hyper-parameter} & \textbf{Value}\\
    \hline
    \# training batches per time step & $2$\\
    RL Batch Size & $1024$ \\
    Discount Factor & $0.99$ \\
    Reward Scaling & $1$ \\
    Replay Buffer Size & $100000$ \\
    Soft Target $\tau$ & $1e-3$ \\
    Target Update Period & $1$ \\
    Use Automatic $\alpha$ tuning & True \\
    Policy Learning Rate & $1e-3$ \\
    Q-function Learning Rate & $1e-3$ \\
    \hline
    \end{tabular}
\caption{SAC training hyper-parameters for all tasks; these are the same values as used in previous work~\cite{skewfit}.}
\label{supptable:sac-parameters}
\end{table*}

%% file: inputs/real-robot-generalization.tex
\section{Generalization to real-world robot}
\label{app:generalization-to-real-world-robot}
\noindent Due to disruptions to lab access caused by the COVID-19 pandemic, we were not able to validate our method on real robots. However, we believe our proposed method can work in the real-world for the following reasons:
\\

\noindent\textbf{ROLL applies simple modifications upon prior work which has been demonstrated to work in the real world.}
Skew-fit, upon which ROLL is based, has been demonstrated on robots in the real-world, for both door opening~\cite{skewfit} and puck pushing~\cite{nair2018visual}. ROLL applies the following modifications upon Skew-Fit, which should not affect the ability of ROLL to work in the real world:

ROLL uses an object-VAE and an LSTM for reward computation. To train the object-VAE, ROLL requires segmented images. As described in section 4.2, the process of unknown object segmentation is fairly simple (see below for a demonstration of our method in the real world). Alternative methods  can also be used for segmentation, such as~\cite{pmlr-v100-xie20b}. 

    Additionally, ROLL needs to pre-collect two datasets for training the scene- and object-VAE. The data collection for the scene-VAE is the same as in prior work~\cite{nair2018visual}. It is also easy to collect the dataset for pre-training the object-VAE: we just need to manually place the target objects at different locations in the scene. We can further reduce the number of manual placements of the object by using data augmentation such as translation, rotation and cropping to create more training samples.

\noindent\textbf{ROLL is more sample-efficient than prior work which has been demonstrated to work in the real world.}
As shown in Figure 4 in the paper, ROLL is much more sample-efficient than Skew-Fit in 4 out of the 5 simulated tasks (and similar efficiency in the 5th).
Sample efficiency is often a major bottleneck for robot learning in the real world. 
As Skew-Fit works on real robots and ROLL is much more sample-efficient, we believe that it will be even easier to apply ROLL to real-world robots.

\noindent\textbf{ROLL consistently works well on 5 different simulated tasks.}
We have tested ROLL in 5 different simulation tasks (using the well-known MuJoCo~\cite{todorov2012mujoco} physics engine), which involves manipulating different objects (pushing a puck, opening a door, and picking up a ball), and with different environments (puck pushing without hurdles and with different kinds of hurdles). The diversity of these tasks provides encouragement that ROLL will work well on similar tasks in the real world (as also demonstrated by the real-world experiments in the previous work that we build on~\cite{skewfit,nair2018visual}).

\begin{figure}[h]
    \centering
    \begin{tabular}{ccccc}
          \includegraphics[width=0.15\textwidth]{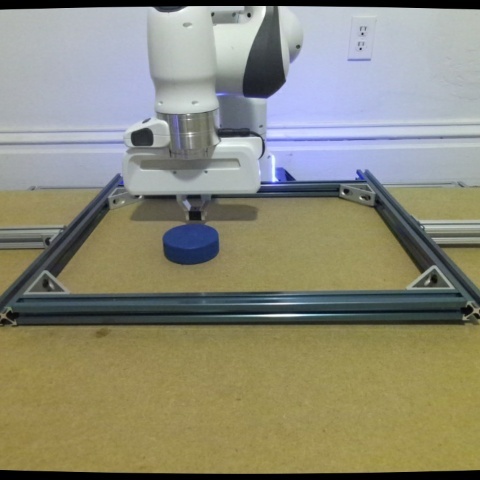} &
          \includegraphics[width=0.15\textwidth]{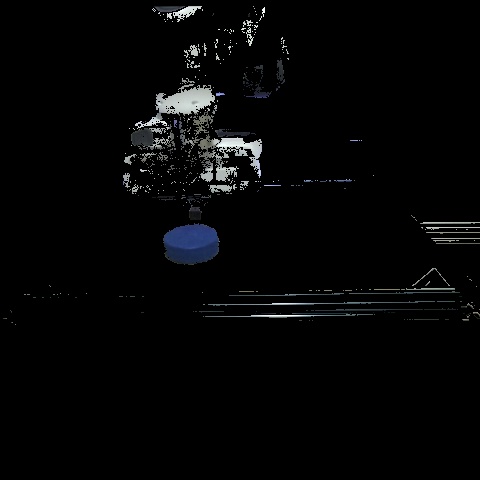} &
          \includegraphics[width=0.15\textwidth]{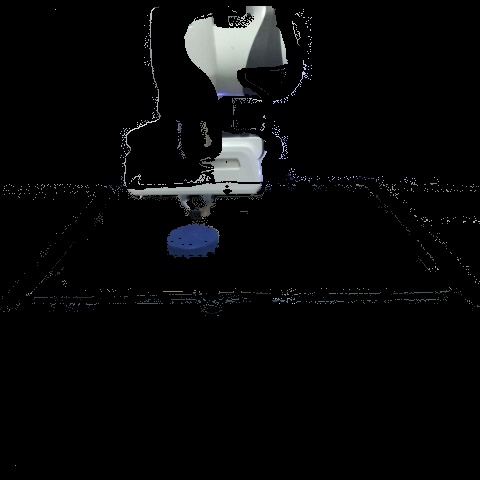} &
          \includegraphics[width=0.15\textwidth]{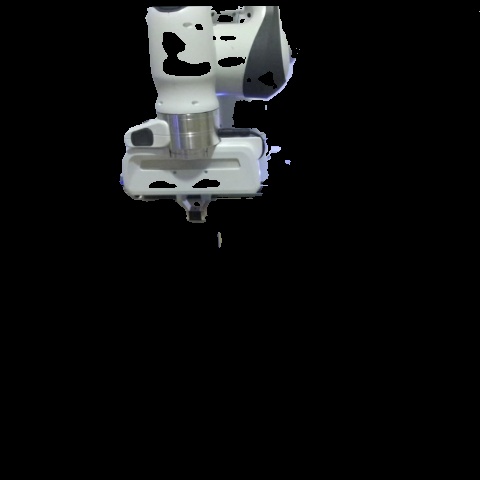} &
          \includegraphics[width=0.15\textwidth]{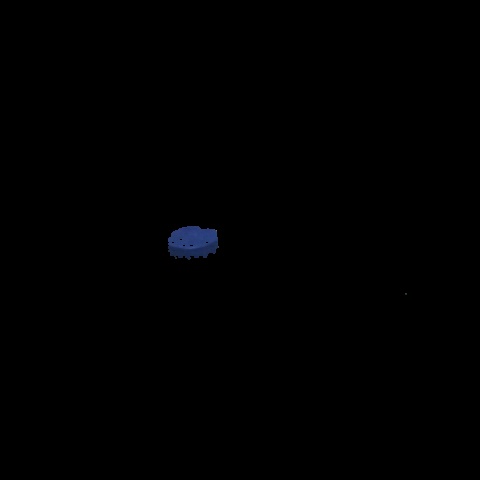} \\
              (a) & (b) & (c) & (d) & (e) 
\end{tabular}
    \caption{Segmentation pipeline in the real world: (a) RGB Input (b) RGB background subtraction 
    (c) Depth background subtraction  (d) Predicted robot mask  (e) Object segmentation. }
    \label{suppfig:real_world_segmentation_pipeline}
\end{figure}

\noindent\textbf{Unknown object segmentation works in the real world.}
To enable our background subtraction method to work robustly in the real wold under shadows and reflections, we train both an RGB and depth-based background subtraction module using \texttt{BackgroundSubtractorMOG2} in OpenCV (Supp Fig.~\ref{suppfig:real_world_segmentation_pipeline}b,c). Similar to our approach in simulation, we train a segmentation network to remove the robot from the foreground (Supp Fig.~\ref{suppfig:real_world_segmentation_pipeline}d).  After removing the background and the robot, what remains is a segmentation of the object (Supp Fig.~\ref{suppfig:real_world_segmentation_pipeline}e).

